\documentclass{article}

\usepackage[preprint]{corl_2026} 

\usepackage{amsmath,amssymb,booktabs,array,graphicx,enumitem,float,wrapfig}
\usepackage{algorithm}
\usepackage{algpseudocode}
\usepackage[font=small]{caption}
\usepackage{cleveref} 
\newcommand{\score}[2]{#1\,{$\pm$\,#2}}
\newcommand{\diagnosticpanel}[1]{%
    \fbox{\begin{minipage}[c][0.50in][c]{0.29\linewidth}\centering\scriptsize #1\end{minipage}}%
}
\title{TRACE: Trajectory-Routed Causal Memory for Delayed-Evidence Visuomotor Imitation}

\author{
  Zihao Li\textsuperscript{1} \quad
  Ranpeng Qiu\textsuperscript{3} \quad
  Yincong Chen \quad
  Guoqiang Ren\textsuperscript{1} \quad
  Weiming Zhi\textsuperscript{2} \\
  \normalfont\textsuperscript{1}Zhejiang University \quad
  \textsuperscript{2}The University of Sydney \\
  \textsuperscript{3}Zhejiang University of Technology
}
\begin{document}
\maketitle

\begin{abstract}
Robots under autonomous operation may require decisions based on evidence that is no longer visible. We study \emph{delayed-evidence} tasks, where an early cue disappears before a later decision point, so visually similar observations can require different actions. In these settings, the current observation is not a sufficient state for control. We introduce TRAjectory-routed Causal Evidence (TRACE), a memory framework for visuomotor imitation policies. TRACE stores task-relevant visual and robot-state evidence, such as object identity, target choice, or route-dependent state, in a fixed-size latent memory that remains bounded over long episodes. Instead of indexing memory by raw time or manually provided task labels, TRACE uses \emph{path signatures}: compact, order-sensitive features of the executed robot-state trajectory. These signatures do not store the visual cue itself; rather, they provide trajectory-conditioned keys for writing and retrieving the evidence stored when the cue was visible. When the robot later reaches an ambiguous observation, the policy conditions on TRACE memory to recover the missing context and choose the correct branch. TRACE attaches through lightweight adapters to policies, without changing the policy backbone, action head, or imitation objective. Across real-world long-horizon manipulation tasks with visually ambiguous branch points, TRACE improves branch selection and task success over alternative baselines, including short-history and recurrent memory. Project page: \url{https://jeong-zju.github.io/trace}
\end{abstract}

\section{Introduction}

Robots often need to make later task decisions using information that is no longer observed. We call these decisions \emph{branches}: alternatives such as which target, route, or manipulation routine to execute next. We study \emph{delayed-evidence manipulation}, where an early cue is observed, disappears, and is only needed later at a decision point. At that point, observations from different histories can look nearly identical while requiring different actions.

Imitation learning \cite{ravichandar2020recent, Diff_templates} for visuomotor policies enables costs that cannot be directly be specified for motion planning. Most visuomotor policies condition on the current observation, possibly with a short recent window, and therefore treat this input as sufficient for action selection. This is an implicit \emph{Markov assumption}: the current input contains all information needed to choose the next action. Delayed-evidence tasks violate this assumption because the present observation does not determine the correct branch. Adding more history is not enough. Short windows fail once the decisive cue falls outside the window, while long windows increase cost and force the policy to infer which earlier observations still matter. Recurrent policies and generic memories can carry information forward, but branch-relevant cues may be overwritten, diluted, or entangled with task-progress signals unrelated to the later choice. Delayed-evidence manipulation therefore requires a causal, fixed-budget memory that is updated online and preserves early task evidence until it becomes useful.

We introduce TRAjectory-routed Causal Evidence (TRACE), a memory framework for visuomotor policies whose current observations are insufficient for action selection. TRACE stores task-relevant visual and robot-state evidence in a fixed set of latent memory slots and updates them online as the robot acts. Rather than indexing memory by raw time or task labels, TRACE uses \emph{path signatures}~\cite{chevyrev2025primer}, which are fixed-length, order-sensitive summaries of the executed robot-state trajectory, as keys for writing and reading memory. These signatures do not store the visual cue itself; instead, they provide trajectory-conditioned access to evidence stored when the cue was visible. This lets TRACE retrieve different stored evidence at visually similar branch points reached through different histories. The policy then conditions on TRACE memory to recover missing context and select the correct branch. TRACE attaches to policies through lightweight adapters, without changing their backbone, action head, or imitation loss.

Concretely, our technical contributions are:
\begin{itemize}[leftmargin=*,nosep]
    \item \textbf{Delayed-evidence imitation:} We formulate a class of visuomotor imitation problems where visually similar branch points require different actions because the decisive evidence appeared earlier in the episode.

    \item \textbf{TRACE memory:} We introduce a fixed-budget causal memory that stores visual and robot-state evidence online, and uses path signatures of the executed robot-state trajectory as order-sensitive keys for writing and reading memory.

    \item \textbf{Plug-in integration and evaluation:} We attach TRACE to action-chunking and diffusion policies through lightweight adapters, without changing their backbone, action head, or imitation loss, and evaluate it on real-world delayed-evidence manipulation tasks with diagnostics showing gains from preserved historical evidence.
\end{itemize}

\begin{figure}[t]
    \centering
    \includegraphics[width=\linewidth]{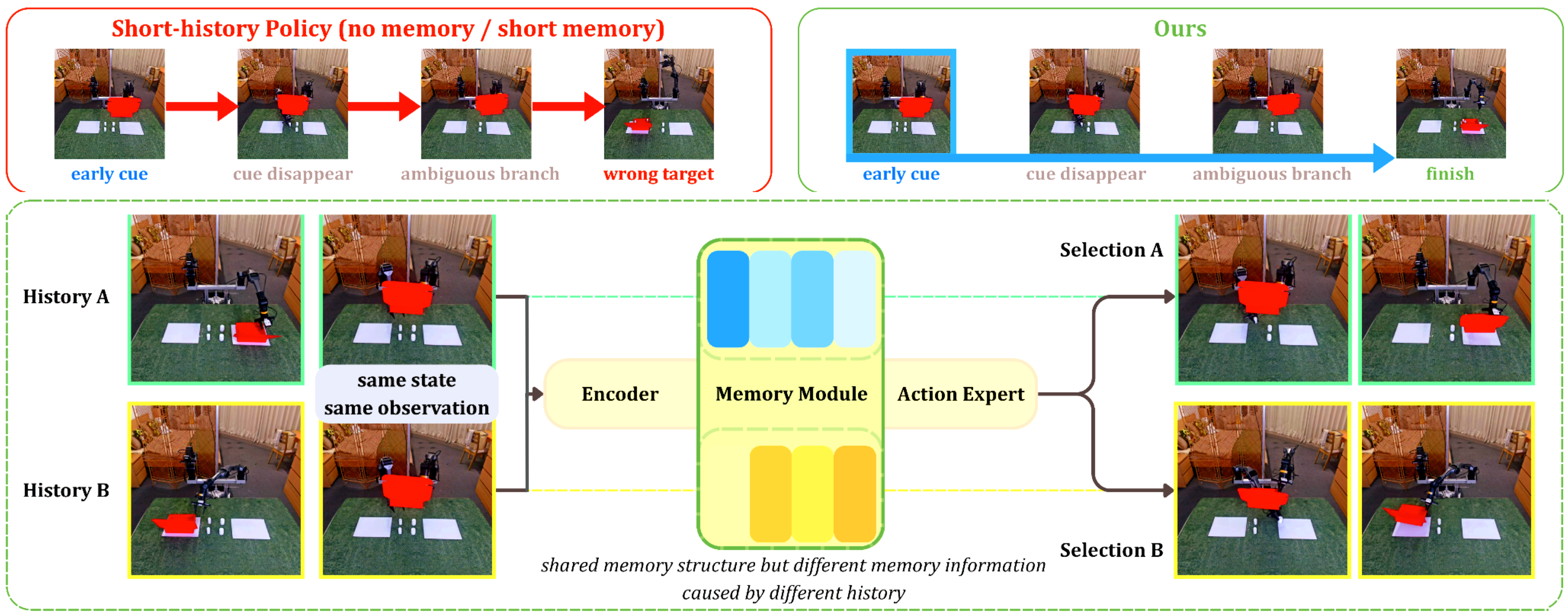}
    \vspace{-1em}
    \caption{\textbf{Delayed evidence in long-horizon manipulation:}
    At a branch point, the robot must choose one task continuation. Observations can look similar even though they require different actions, based on the past. A short-history policy fails because its window contains the latest information but not any historical cues. TRACE stores the cue when it is visible and reads that memory later to enable correct selection.}
    \label{fig:teaser}
\end{figure}


\section{Related Work}

\textbf{Visuomotor imitation policies and memory:}
Modern visuomotor policies map recent observations to actions, action chunks, or action distributions. Action-chunking policies predict short horizons with temporal aggregation~\citep{zhao2023learning}, probabilistic representations of motion enable reasoning over uncertainty \cite{zhi2024diagrammatic, promp}, while diffusion policies generate action sequences through iterative denoising~\citep{chi2025diffusion}. Dynamical system-based policies are another representation that allows for guarantees of robustness \cite{Fast_diff_int, periodic}. Larger generalist policies extend this paradigm with language conditioning and broad robot datasets~\citep{brohan2022rt1,zitkovich2023rt,o2024open,black2024pi0,shukor2025smolvla,zheng2025xvla,nvidia2025gr00tn1}. Historically, memory about the environment is encapsulated in map representations of the environment \cite{HM, sptemp}. Despite these differences, their behaviour is limited by the information available in the conditioning state. In delayed-evidence tasks, the current frame or short observation window may no longer contain the cue needed for a later branch. Recurrent policies, context windows, and explicit memory mechanisms address partial observability by carrying information forward, but can require backbone changes, long contexts, demonstration retrieval, or planner-specific memories~\citep{cherepanov2025elmur,li2025map,lin2025echovla}. TRACE instead attaches a fixed-budget online memory to the action policy through lightweight adapters, without changing the policy backbone, action head, or imitation objective.

\textbf{Trajectory signatures and latent history representations:}
Path signatures provide compact, order-sensitive summaries of continuous trajectories and have been used as sequence features and recurrent gating mechanisms~\citep{chevyrev2025primer,kidger2020signatory}. Recurrent policies, transformer context windows, and explicit memory systems address partial observability by carrying information forward. Recent examples include layer-local external memory~\citep{cherepanov2025elmur}, retrieval-prompt memories for frozen policies~\citep{li2025map}, declarative scene and episodic memories~\citep{lin2025echovla}, and keyframe memories for hierarchical imitation~\citep{buamanee2026bi}. These methods improve temporal context, but often require modifying the policy backbone, retrieving from demonstrations, or specialising memory for a planner. TRACE uses latent history differently: path and delta signatures are not a policy backbone, recurrent hidden state, or future-prediction objective. They serve as trajectory-conditioned keys for writing and reading visual and robot-state evidence in a fixed-budget memory, linking memory access to the executed trajectory while leaving the base imitation loss unchanged.


\section{Problem Setup and Background}
\label{sec:setup_notation}
We use \emph{history} to mean the causal execution information available before selecting the next action.

\textbf{Delayed-evidence imitation:}
We study imitation tasks with an early cue, a shared execution segment, and a later ambiguous branch point. A \emph{branch} is one possible task continuation, such as a target, route, or manipulation routine, and the \emph{branch point} is the time when the policy must choose among these continuations. At this point, different histories can produce visually similar observations but require different expert actions, so the ambiguity cannot be resolved from the current observation alone.

We consider demonstrations $\mathcal{D}=\{\tau_i\}_{i=1}^{N}$, where $\tau_i=\{(o_t^i,s_t^i,a_t^i)\}_{t=1}^{T_i}$ contains a multi-view observation $o_t$, robot state $s_t$, and demonstrated action $a_t$. Let $x_t=(o_t,s_t)$ and let $\mathcal{H}_t=(x_1,a_1,\ldots,x_{t-1},a_{t-1},x_t)$ denote the history available before choosing $a_t$. In delayed-evidence tasks, $\mathcal{H}_t$ may contain task-relevant evidence that is absent from the current observation window. Thus, for a short window of length $w$, there may be histories $\mathcal{H}_t$ and $\mathcal{H}'_t$ such that
\begin{align}
    x_{t-w+1:t} \approx x'_{t-w+1:t},
    \qquad
    a_t^\star(\mathcal{H}_t) \neq a_t^\star(\mathcal{H}'_t),
\end{align}
where $a_t^\star(\mathcal{H}_t)$ denotes the expert target required after history $\mathcal{H}_t$. We use the single-step action notation for the problem setup, while the policy-native prediction target in our implementations may be an action, action chunk, action-token sequence, or diffusion denoising target.

\textbf{Path signatures as streaming trajectory keys:}
TRACE needs a causal memory key: a compact descriptor of how the robot reached the current state, maintained online without storing the full history. Given the piecewise-linear interpolation of the robot-state trajectory up to time $t$, written as $S_t:[0,1]\rightarrow\mathbb{R}^{d_s}$, its depth-$p$ path signature is
\begin{align}
    \operatorname{Sig}_{\leq p}(S_t)
    =
    \left(
    1,\,
    \int dS,\,
    \int_{u_1<u_2} dS_{u_1}\otimes dS_{u_2},\,
    \ldots,\,
    \int_{u_1<\cdots<u_p} dS_{u_1}\otimes\cdots\otimes dS_{u_p}
    \right).
    \label{eq:path_signature_setup}
\end{align}
Signatures summarise both net changes and ordered interactions among state-coordinate changes, making them sensitive to how a trajectory unfolds rather than only which states it visits~\citep{kidger2020signatory}. For continuous paths, they are invariant to monotone time reparameterisation; for sampled robot trajectories, the piecewise-linear signature gives a descriptor that is typically less tied to raw execution speed than time-step indexing. This gives TRACE a deterministic, fixed-budget trajectory key, rather than using another learned recurrent state as both memory and address. Signatures also support streaming updates, so TRACE can maintain the trajectory key online as new states arrive.

In TRACE, signatures are deterministic trajectory descriptors used for memory access. They are not task labels, visual memories, recurrent hidden states, or policy outputs. Visual and robot-state features store the task evidence, while signatures help determine where that evidence is written and read. We denote the streamed trajectory signature by $\xi_t=\operatorname{Sig}_{\leq p}(S_t)$ and and a local change feature in signature coordinates, $\delta_t=\xi_t-\xi_{t-1}$, with $\delta_1=\mathbf{0}$. We give signature dimension details in Appendix~\ref{app:signature_depth} and the robot-state representation in Appendix~\ref{app:state_representation}.


\begin{figure}[t]
    \centering
    \includegraphics[width=\linewidth]{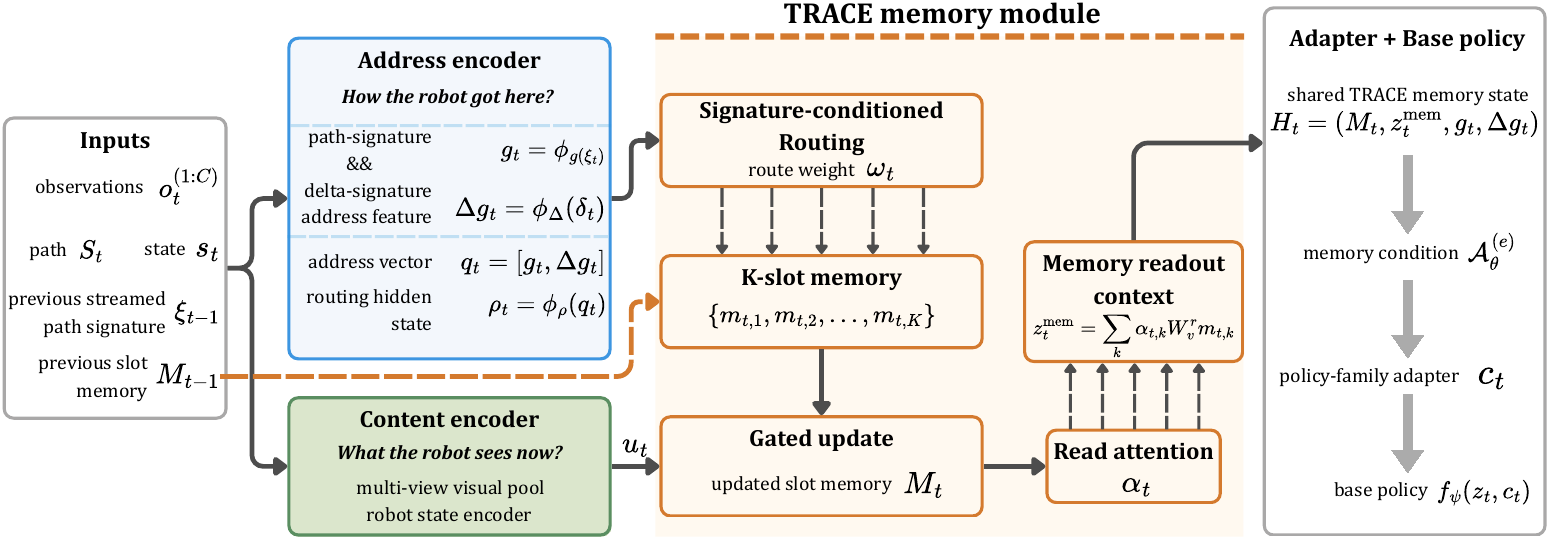}
    \caption{\textbf{TRACE signal flow.}
    TRACE encodes current visual-state evidence as memory content, uses streamed path-signature features as trajectory-derived keys, updates fixed-size latent memory slots, and returns a compact memory condition to the base visuomotor policy.}
    \label{fig:signal_diagram}
\end{figure}


\section{TRAjectory-routed Causal Evidence (TRACE)}
\label{sec:method}

TRACE augments an existing visuomotor imitation policy with a fixed-size causal memory. During execution, it writes task-relevant visual and robot-state evidence into latent slots, and later reads this memory when the current observation becomes ambiguous. The base policy keeps its original action representation, action head, and imitation loss; TRACE adds only a memory-conditioning pathway through a lightweight adapter. Figure~\ref{fig:signal_diagram} summarises the components of TRACE.

\subsection{Causal Memory Interface and Trajectory Keys}
\label{sec:method_memory_interface}

Let $z_t=x_{t-w+1:t}$ be the recent observation window used by the base policy, where $x_t=(o_t,s_t)$ contains multi-view observations $o_t$ and robot state $s_t$. We write the downstream policy as
\begin{align}
    \hat y_t=f_{\psi}(z_t,c_t),
\end{align}
where $\hat y_t$ is the policy's native prediction target, such as an action, action chunk, action-token sequence, or diffusion denoising target. TRACE supplies $c_t$, a memory condition containing causal information that may no longer be visible in $z_t$.

TRACE separates \emph{memory content} from \emph{memory access}. The current image and robot state provide what should be remembered, while the executed trajectory determines where that information is stored and later retrieved. Thus, signatures are not simply concatenated to the policy input; they route memory writes and reads over a fixed set of latent slots. At each timestep, after observing $x_t$ and before selecting the next action, TRACE writes the current evidence into memory, reads from the updated memory, and then queries the policy:
\begin{align}
    R_t &= (M_t,z_t^{\mathrm{mem}},g_t,\Delta g_t),
    \qquad
    c_t = \mathcal{A}_{\theta}^{(e)}(R_t),
    \qquad
    \hat y_t=f_{\psi}(z_t,c_t).
    \label{eq:trace_interface}
\end{align}
Here $M_t\in\mathbb{R}^{K\times d}$ is a $K$-slot latent memory, $z_t^{\mathrm{mem}}$ is the memory readout, $g_t$ and $\Delta g_t$ are trajectory-derived key features, and $\mathcal{A}_{\theta}^{(e)}$ is the adapter for policy family $e$.

To form the trajectory key, TRACE uses the normalised robot-state path up to time $t$, written as a piecewise-linear path $S_t:[0,1]\rightarrow\mathbb{R}^{d_s}$. It maintains a streamed depth-$p$ path signature $\xi_t=\operatorname{Sig}_{\leq p}(S_t)$ and a signature-space increment $\delta_t=\xi_t-\xi_{t-1}$, with $\delta_1=\mathbf{0}$. The cumulative signature $\xi_t$ provides a global trajectory address, while the increment $\delta_t$ exposes recent motion in the same coordinate system, allowing routing to depend on both long-horizon path context and local progress. Learned projections embed these as $g_t=\phi_g(\xi_t)$ and $\Delta g_t=\phi_{\Delta}(\delta_t)$, and form the trajectory key
\begin{align}
    q_t=\phi_q([g_t,\Delta g_t]).
    \label{eq:trajectory_key}
\end{align}
All $\phi$ maps are learned projections or MLPs unless otherwise stated. The key point is that $q_t$ is trajectory-derived: it indexes memory access using the ordered robot-state history, while the memory contents store visual-state evidence.

\subsection{Signature-Routed Slot Memory}
\label{sec:method_signature}

TRACE uses the trajectory key to route writes and reads over a fixed-size latent memory. At each step, the current visual-state input is encoded as $e_t=\phi_x(x_t)$, a pooled multi-view visual and proprioceptive feature. This feature contains the evidence currently available to the robot, such as an object identity, target choice, or route-dependent cue. TRACE can therefore store a cue when it is visible and expose it later when the current observation no longer contains that information.

The memory contains slots $M_t=\{m_{t,1},\ldots,m_{t,K}\}$, with each slot $m_{t,k}\in\mathbb{R}^{d}$. Slots are reset at the start of each demonstration scan or online episode. Because routing depends on the current trajectory key and the previous slot contents, slot selection can adapt as memory contents evolve rather than following a fixed hash of the trajectory. Given the trajectory key $q_t$, TRACE computes a routing state $\rho_t=\phi_{\rho}(q_t)$ and compares it with the previous slot contents:
\begin{align}
    \bar m_{t-1,k} &= m_{t-1,k}+\lambda_{\eta}\eta_k,
    \qquad
    \ell_{t,k}
    =
    \frac{(W_Q\rho_t)^\top W_K \bar m_{t-1,k}}
    {\tau\sqrt{d_r}},
    \qquad
    \omega_{t,k}=\operatorname{softmax}_k(\ell_{t,k}).
    \label{eq:trace_routing}
\end{align}
Here $\eta_k$ is an optional fixed slot identity, $\lambda_{\eta}$ controls its weight, $\tau$ is a routing temperature, and $\omega_{t,k}$ determines how strongly slot $k$ is selected by the current trajectory key. The slot identities break initial slot symmetry during addressing, but are not stored as recurrent memory content.

TRACE writes the current evidence into selected slots through a gated update. We first form a write input, $u_t = \phi_w(e_t,q_t)$, which combines the current visual-state evidence with the trajectory key. The slot update is then
\begin{align}
    \tilde m_{t,k}
    &=
    \tanh\!\left(\phi_m(\bar m_{t-1,k},u_t,\rho_t)\right),
    \qquad
    \beta_{t,k}
    =
    \omega_{t,k}\,
    \sigma\!\left(\phi_{\beta}(\bar m_{t-1,k},u_t,\rho_t)\right),
    \\
    m_{t,k}
    &=
    (1-\beta_{t,k})m_{t-1,k}
    +
    \beta_{t,k}\tilde m_{t,k}.
    \label{eq:trace_memory_update}
\end{align}
Here $\tilde m_{t,k}$ is the proposed new content for slot $k$, and $\beta_{t,k}$ is the routing-modulated write gate. Slots with small $\omega_{t,k}$ remain nearly unchanged, while selected slots absorb the current visual-state evidence. After writing, TRACE reads from the updated slots using a query formed from the current evidence and trajectory state:
\begin{align}
    \alpha_{t,k}
    =\operatorname{softmax}_k\!\left(
    \frac{\phi_r(e_t,\rho_t)^\top W_K^{r}(m_{t,k}+\lambda_{\eta}\eta_k)}{\sqrt d}
    \right),
    \qquad
    z_t^{\mathrm{mem}}=\sum_{k=1}^{K}\alpha_{t,k}W_V^{r}m_{t,k}.
    \label{eq:trace_memory_read}
\end{align}
Here $\alpha_{t,k}$ is the read weight, and $W_K^{r},W_V^{r}$ are learned read-key and read-value projections. The shared TRACE state is $R_t=(M_t,z_t^{\mathrm{mem}},g_t,\Delta g_t)$. Here $M_t\in\mathbb{R}^{K\times d}$ is the updated $K$-slot latent memory, $z_t^{\mathrm{mem}}$ is the readout from $M_t$, $g_t$ and $\Delta g_t$ are trajectory-derived key features, and $\mathcal{A}_{\theta}^{(e)}$ is the adapter for policy family $e$. Full projection, auxiliary-loss, and pseudocode details are given in Appendix~\ref{app:memory_update_losses}.

Both training and deployment use the same causal scan: TRACE receives only observations and robot states available before the policy query, and does not use cue labels, branch labels, future frames, or test-time demonstration retrieval. At each step, the memory update uses only observations and states available up to that time, and the resulting condition $c_t$ is passed to the unchanged imitation loss of the base policy.

\subsection{Policy Adapter: From TRACE Memory to Policy Conditioning}
\label{sec:method_adapter}

The TRACE updater is shared across policy families; only the policy-facing adapter changes. Given the shared TRACE state $R_t=(M_t,z_t^{\mathrm{mem}},g_t,\Delta g_t)$, the adapter maps memory into the native conditioning space of policy family $e$ as $c_t^{(e)}=\mathcal{A}_{\theta}^{(e)}(R_t)\in\mathcal{C}_e$. The base policy's action representation, action head, and supervised imitation loss are unchanged.

For action-chunking regression policies, the adapter projects the memory slots into 512-D memory tokens and maps $[z_t^{\mathrm{mem}},g_t,\Delta g_t]$ into a summary token. These tokens are concatenated to the policy's attention memory, after which the original action head regresses the native action chunk. For diffusion policies, the adapter pools the slots using the read weights, concatenates the pooled memory with $z_t^{\mathrm{mem}}$, $g_t$, and $\Delta g_t$, and maps the result through a zero-initialised MLP into an additive global conditioning vector. The denoising process and diffusion loss are otherwise unchanged.

Thus, the adapter is only a translator from TRACE memory to the policy's existing conditioning interface, not a new policy backbone, action decoder, or retrieval module. Appendix~\ref{app:adapter_architectures} gives architectural details, parameter counts, latency, and training details.

\section{Empirical Evaluations}
\label{sec:result}

We organise the experiments around four questions: (1) Does adding causal memory improve delayed-evidence manipulation? (2) Does the gain transfer across policy families? (3) Does TRACE improve over generic history memory? (4) Does TRACE use historical evidence rather than relying only on the decision-point observation? We answer these questions through base-policy comparisons, cross-policy TRACE variants, memory ablations, and history-transformation diagnostics.

\begin{wrapfigure}{r}{0.45\linewidth}
    \centering
    \includegraphics[width=\linewidth]{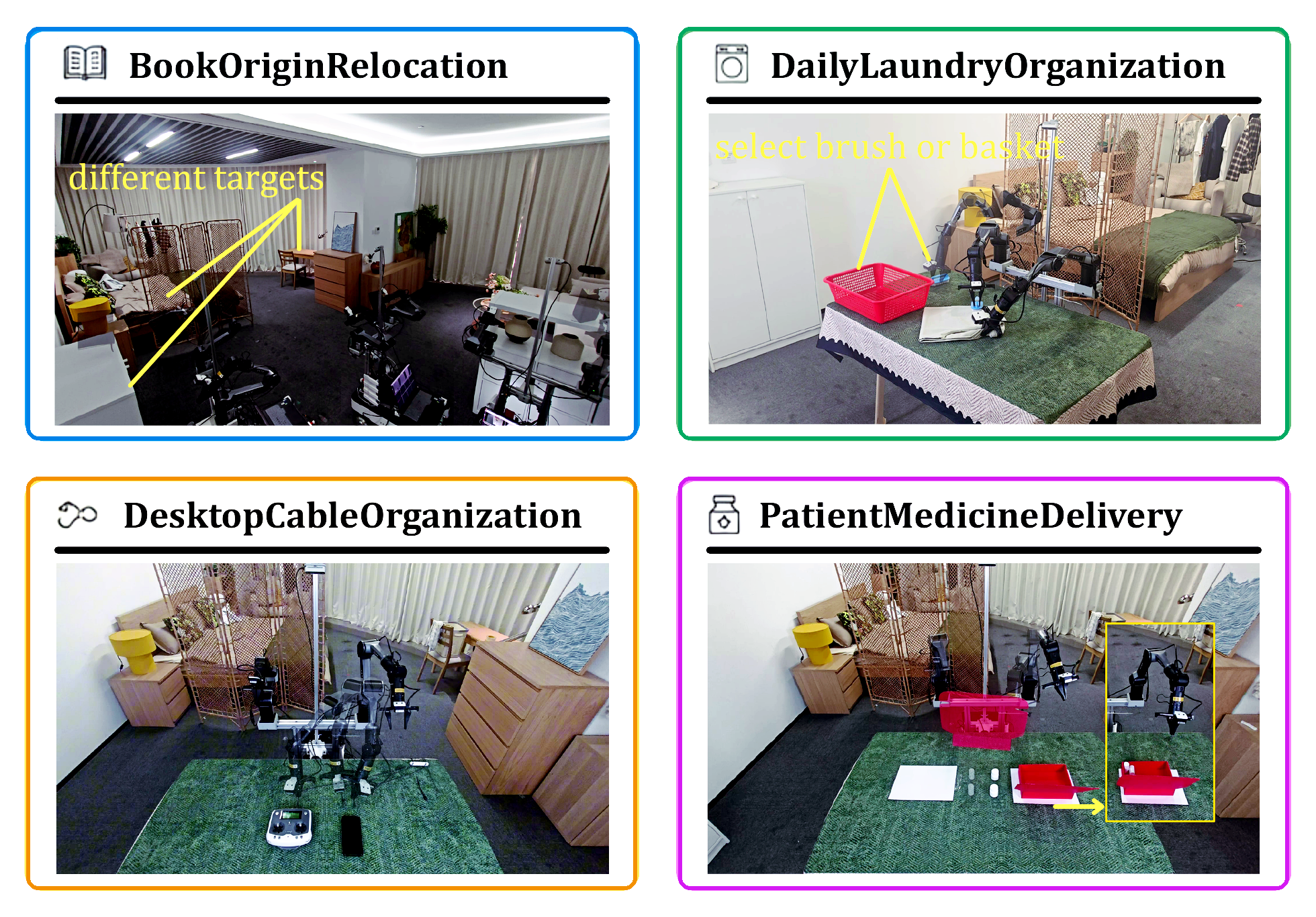}
    \caption{Overview of the selected delayed-evidence manipulation tasks.}
    \label{fig:tasks_overview}
    \vspace{-0.8em}
\end{wrapfigure}

\textbf{Experimental setup.}
Figure~\ref{fig:tasks_overview} illustrates the delayed-evidence task suite. We collect data with the system described in \cite{li2026tripilot}. We evaluate on five real-world tasks: \emph{Tool}, where initial object identity determines a later tool sequence; \emph{Book}, where origin determines route and placement; \emph{Laundry}, where origin side determines brush-and-basket versus fold-and-store; \emph{Cable}, where origin side determines the matched device; and \emph{Medicine}, where tray origin determines box selection and return. Appendix~\ref{app:dataset_demonstration_details} gives task, dataset, and rollout details.

\textbf{Baselines.}
We attach TRACE to both regression-style action-chunking policies and diffusion-based policies, and compare against the corresponding base policies. We also evaluate deployable robot policy families used for imitation or language-conditioned control, including SmolVLA~\citep{shukor2025smolvla}, $\pi_{0.5}$~\citep{black2024pi0}, X-VLA~\citep{zheng2025xvla}, and GR00T N1.6~\citep{nvidia2025gr00tn1}. These VLA baselines are not frozen zero-shot models: each is adapted on the same single-task demonstrations and train/evaluation split. Appendix~\ref{app:baseline_protocols} gives the model setups, and Appendix~\ref{app:model_size_hparams} reports training hyperparameters and parameter counts for all baselines and TRACE modules.

\textbf{Metrics.}
Following partial-progress reporting in long-horizon manipulation benchmarks~\citep{heo2025furniturebench,mees2022calvin}, we score each rollout by stage-level progress over ordered subtasks, including manipulation stages and memory-dependent branch decisions. We report task-balanced averages, rollout standard errors, bootstrap uncertainty, and full scoring details in Appendix~\ref{app:problem_setup_details}.

\textbf{Question 1. Does memory help delayed-evidence manipulation?}
Yes. Table~\ref{tab:main_results} compares base visuomotor policies, fine-tuned VLA-style baselines, and the same base policy families augmented with TRACE causal memory. TRACE improves both policy families substantially: Regression rises from $25.50$ to $69.23$ average progress, and Diffusion rises from $25.00$ to $59.53$. TRACE also outperforms the strongest non-TRACE baseline, $\pi_{0.5}$, by $18.76$ points for Regression and $9.06$ points for Diffusion. The largest gains occur on tasks where an origin or object cue observed early determines a later branch after the cue has left view.

\begin{table}[t]
    \centering
    \scriptsize
    \caption{Main comparison on real-world delayed-evidence tasks. Values are mean stage progress (\%) $\pm$ SE.}
    \label{tab:main_results}
    \renewcommand{\arraystretch}{1.05}
    \resizebox{\linewidth}{!}{%
    \begin{tabular}{@{}lcccccc@{}}
\toprule
Method                   & Tool$\uparrow$       & Book$\uparrow$      & Laundry$\uparrow$   & Cable$\uparrow$      & Medicine$\uparrow$  & Avg. progress$\uparrow$ \\
\midrule
Diffusion Policy         & \score{18.00}{1.41}  & \score{12.33}{0.44} & \score{22.00}{0.32} & \score{34.00}{0.48}  & \score{38.67}{1.01} & \score{25.00}{0.38} \\
ACT                      & \score{31.00}{5.51}  & \score{13.67}{0.49} & \score{11.50}{2.01} & \score{44.00}{7.79}  & \score{27.33}{0.05} & \score{25.50}{1.95} \\
$\pi_{0.5}$              & \score{45.50}{0.81}  & \score{51.00}{6.99} & \score{47.50}{1.45} & \score{49.00}{0.57}  & \score{59.33}{1.45} & \score{50.47}{1.47} \\
GR00T N1.6               & \score{39.00}{2.31}  & \score{12.67}{2.14} & \score{24.00}{2.72} & \score{38.00}{12.11} & \score{39.33}{3.29} & \score{30.60}{2.64} \\
SmolVLA                  & \score{25.00}{1.51}  & \score{20.00}{0.82} & \score{12.00}{2.16} & \score{50.00}{1.44}  & \score{34.67}{1.01} & \score{28.33}{0.65} \\
X-VLA                    & \score{5.50}{0.94}   & \score{47.00}{6.19} & \score{41.00}{0.36} & \score{36.00}{21.44} & \score{40.00}{2.92} & \score{33.90}{4.51} \\
\midrule
\textbf{Ours (Regression)}     & \score{54.50}{17.96} & \score{83.00}{0.86} & \score{81.00}{2.84} & \score{51.00}{1.11}  & \score{76.67}{1.05} & \score{69.23}{3.65} \\
\textbf{Ours (Diffusion)} & \score{58.50}{10.17} & \score{68.33}{2.86} & \score{70.50}{0.76} & \score{51.00}{4.63}  & \score{49.33}{0.17} & \score{59.53}{2.31} \\
\bottomrule
    \end{tabular}}
\end{table}

\textbf{Takeaway.}
TRACE improves delayed-evidence manipulation over policies without causal memory, with the clearest gains on Book, Laundry, and Medicine, where an early origin cue determines a later visually ambiguous branch. The aggregate result is not driven solely by the high-variance Tool task: Appendix~\ref{app:variance_failure} reports leave-one-task-out and Tool-specific analyses. Appendix~\ref{app:additional_case_studies} further connects common branch errors, rollout evidence, and TRACE behaviour to the task cases. Deployment overhead is measured in Appendix~\ref{app:runtime_latency}.

\textbf{Question 2. Is the gain confined to a specific policy family?}
No. Table~\ref{tab:main_results} shows that TRACE improves both the Regression and Diffusion variants while using the same updater, the same signature routing, and the same history scan used in training. Only the policy-facing adapter changes: Regression uses the adapter for the action-chunking regressor, while Diffusion uses the adapter for the diffusion policy. Regression is stronger on average in these measured rollouts. Its memory condition enters in a single forward pass, while the diffusion sampler carries that condition through repeated refinement. The important point is that both policy families benefit from the same TRACE memory state. This supports the modular design claim that TRACE supplies missing history information while leaving the downstream imitation objective and action head intact.

\begin{figure}[!t]
    \centering
    \IfFileExists{figs/rollout_results_combined.pdf}{%
        \includegraphics[width=\linewidth]{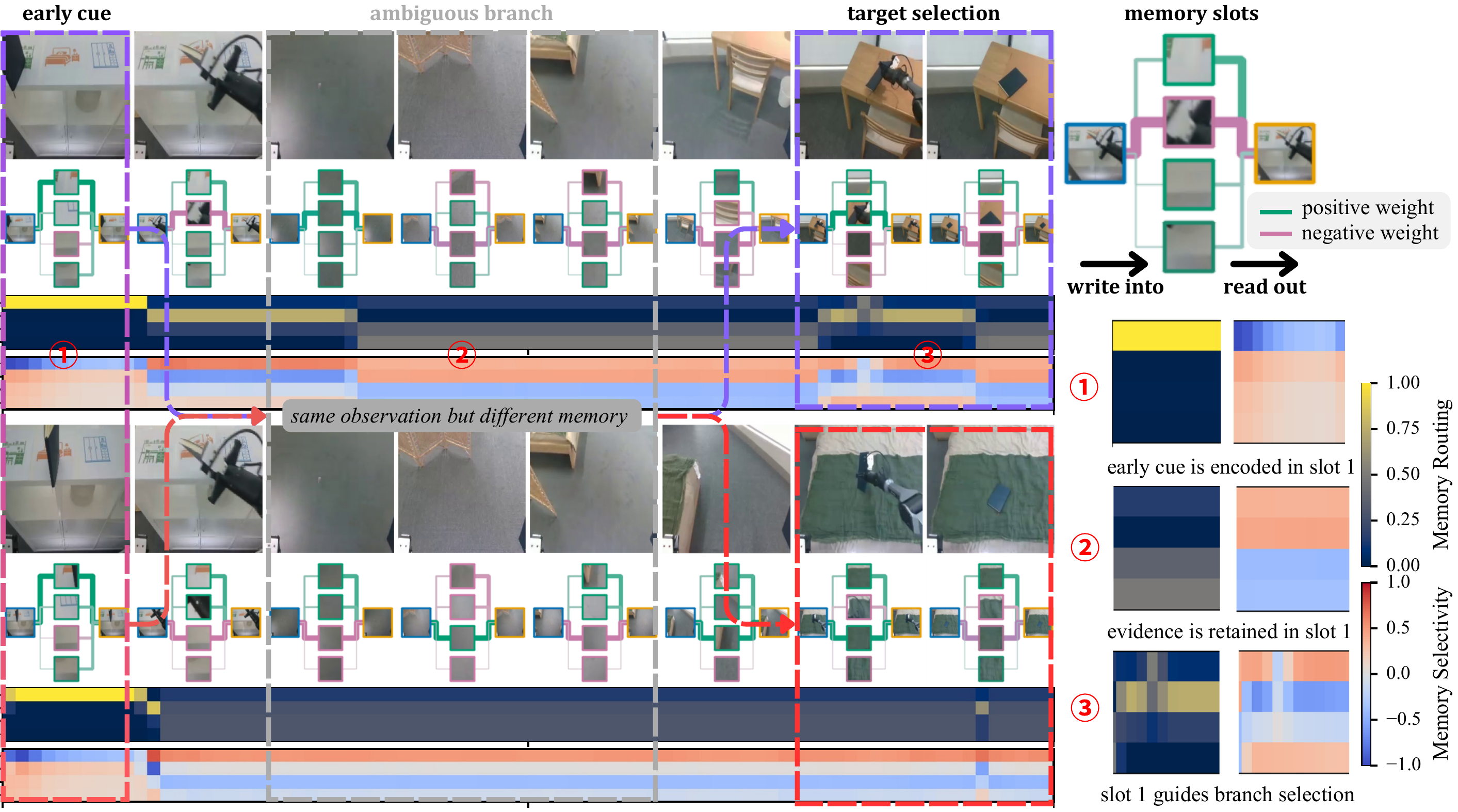}%
    }{%
        \diagnosticpanel{past cue}\hfill
        \diagnosticpanel{ambiguous branch}\hfill
        \diagnosticpanel{target selection}%
    }
    \caption{Rollout results for \emph{Book}. The timeline contains past cue, visually ambiguous transit, and target selection, while the overlaid slot graph and right panels show where evidence is written, retained, and read. Positive and negative denote signed memory weights: positive weights add support for the selected slot, whereas negative weights carry opposite-sign evidence that suppresses these slots.}
    \label{fig:rollout_results_combined}
\end{figure}

Figure~\ref{fig:rollout_results_combined} visualises the measured memory signals. Signature-indexed writes do not collapse to one slot. Routing changes over episode progress, while the readout remains selective for slots carrying history evidence near the branch point. In the Book rollout, the overhead view becomes visually similar after pickup and transit, but the memory graph preserves the origin-dependent trajectory through signed signature scores and high-weight slot edges. Appendix~\ref{app:additional_case_studies} gives additional task-level case studies that connect these diagnostics to concrete rollout failures and recoveries.

\textbf{Question 3. Is TRACE better than generic history memory?}
Yes, when the delayed cue must be preserved through the executed causal history. Table~\ref{tab:memory_module_results} compares TRACE with recurrent, history-token, external-memory, and retrieval alternatives under the same regression base policy, normalisation, and evaluation blocks. These controls ask whether delayed-evidence manipulation only needs more history, or whether the history must be organised around the executed causal history.

\begin{table}[!t]
    \centering
    \caption{Memory-module comparison. The recurrent, transformer-context, LRU, and retrieval-prompt controls are inspired by GRU gating~\citep{cho2014learning}, self-attention context~\citep{vaswani2017attention}, least-recently-used external memory access~\citep{santoro2016meta}, and memory-augmented prompting~\citep{li2025map}. Values are mean stage progress percentages $\pm$ SE.}
    \label{tab:memory_module_results}
    \setlength{\tabcolsep}{2.2pt}
    \renewcommand{\arraystretch}{0.9}
    \resizebox{\linewidth}{!}{%
    \tiny
    \begin{tabular}{@{}lccccccccc@{}}
        \toprule
        Memory module & Online & Fixed & Sig. & Tool & Book & Laundry & Cable & Medicine & Avg. \\
        \midrule
        No memory & -- & -- & -- & \score{31.00}{5.51} & \score{13.67}{0.49} & \score{11.50}{2.01} & \score{44.00}{7.79} & \score{27.33}{0.05} & \score{25.50}{1.95} \\
        GRU recurrent memory & \checkmark & \checkmark & -- & \score{40.50}{9.12} & \score{49.00}{3.04} & \score{44.00}{2.21} & \score{47.00}{4.98} & \score{48.67}{1.24} & \score{45.83}{1.91} \\
        Transformer history context & -- & -- & -- & \score{40.50}{7.83} & \score{61.67}{2.54} & \score{55.00}{1.96} & \score{46.00}{3.87} & \score{57.33}{1.41} & \score{52.10}{1.68} \\
        LRU external memory & \checkmark & \checkmark & -- & \score{46.50}{7.64} & \score{57.67}{2.48} & \score{54.50}{2.03} & \score{51.00}{3.52} & \score{60.00}{1.22} & \score{53.93}{1.61} \\
        Retrieval-prompt memory & -- & -- & -- & \score{48.00}{7.66} & \score{62.67}{2.18} & \score{59.50}{1.77} & \score{50.00}{3.62} & \score{61.33}{1.12} & \score{56.30}{1.46} \\
        \textbf{TRACE signature-routed slots} & \checkmark & \checkmark & \checkmark & \score{54.50}{17.96} & \score{83.00}{0.86} & \score{81.00}{2.84} & \score{51.00}{1.11} & \score{76.67}{1.05} & \score{69.23}{3.65} \\
        \bottomrule
    \end{tabular}}
\end{table}

\textbf{Takeaway.} Generic memory helps relative to no memory, but Table~\ref{tab:memory_module_results} shows that context length or storage capacity alone does not match TRACE. The matched controls can retain history, but they do not explicitly bind the delayed cue to the executed trajectory address that will be read at the branch point. Appendix~\ref{app:additional_case_studies} connects this distinction to task-level branch errors and TRACE recoveries.

\textbf{Question 4. Do diagnostics show that TRACE uses earlier evidence rather than only the observation at the decision point?}
Yes. Table~\ref{tab:diagnostic_results} reports ablations and history-transform diagnostics that test whether TRACE uses the executed history. Route similarity measures whether a transformed online rollout history writes through the same slot route as the original TRACE rollout before the branch point. Branch consistency measures whether the branch-point readout still carries the same delayed branch choice. Time resampling, speed jitter, state offset, and sparse sampling should preserve the relevant history evidence, while order reversal deliberately destroys the causal order of the history. We normalise each raw diagnostic by the corresponding Full TRACE value on the same task for cross-task comparability. Appendix~\ref{app:diagnostic_metrics} gives the full formulas and averaging.

\begin{table}[!t]
    \centering
    \caption{Ablations and history-transform diagnostics. Route similarity measures agreement between the transformed online rollout history and the original Full TRACE rollout's slot routing sequence. Branch consistency measures agreement between the branch point memory readout and the branch action. Both diagnostics are normalised to Full TRACE only for cross-task comparison.}
    \label{tab:diagnostic_results}
    \setlength{\tabcolsep}{2.2pt}
    \renewcommand{\arraystretch}{0.9}
    \resizebox{\linewidth}{!}{%
    \tiny
    \begin{tabular}{@{}lc@{\hspace{10pt}}lccc@{}}
        \toprule
        \multicolumn{2}{c}{Component ablations} & \multicolumn{4}{c}{History-transform diagnostics} \\
        \cmidrule(r){1-2}\cmidrule(l){3-6}
        Variant & Avg.$\uparrow$ &
        History transform & Tested property & Route similarity$\uparrow$ & Branch consistency$\uparrow$ \\
        \midrule
        Current observation only & \score{25.50}{1.95} &
        Time resampling & time-change inv. & \score{92.40}{1.10} & \score{96.00}{0.80} \\
        Signature-only & \score{45.50}{1.82} &
        Speed jitter & time-change inv. & \score{89.80}{1.40} & \score{94.70}{1.00} \\
        Unrouted slot memory & \score{52.17}{1.63} &
        State offset & offset inv. & \score{91.20}{1.20} & \score{95.10}{0.90} \\
        No-delta routing & \score{61.43}{2.21} &
        Sparse sampling & sampling robust. & \score{86.50}{1.60} & \score{92.30}{1.20} \\
        Mean readout & \score{62.80}{2.44} &
        Order reversal & negative control & \score{37.80}{2.50} & \score{41.60}{2.20} \\
        No auxiliary losses & \score{66.10}{2.84} &
        Order-preserving controls & summary & \score{89.98}{0.68} & \score{94.53}{0.49} \\
        \textbf{Full TRACE} & \score{69.23}{3.65} &
        \textbf{Full TRACE} & reference & \score{100.00}{0.00} & \score{100.00}{0.00} \\
        \bottomrule
    \end{tabular}}
\end{table}

\textbf{Takeaway.} Table~\ref{tab:diagnostic_results} supports the intended behaviour. The order-preserving transforms keep route similarity and branch consistency close to the Full TRACE reference, while order reversal sharply lowers both diagnostics. This contrast ties the memory readout to ordered historical evidence rather than to the unchanged observation at the decision point.

\section{Limitations and Conclusion}
\label{sec:conclusion}
\textbf{Limitations:} The main limitation is the dimensionality of the signature. For standard truncated signatures, the raw feature dimension grows rapidly with the robot-state dimension and signature depth before learned compression. This can increase normalisation, projection, memory, and computation costs, and may make TRACE less efficient when higher-dimensional states or deeper signatures are required.

\textbf{Conclusion:} TRACE gives visuomotor robot policies a compact causal memory state for the executed history. By routing visual-state writes with path and delta signatures, it stores past task evidence in a fixed-size TRACE memory state and presents that memory through a lightweight adapter. This yields a modular design where memory and action generation remain separate from the base policy objective. Across five real-world delayed-evidence tasks, the same memory module improves both regression and diffusion policy families, and diagnostics show that the gains come from remembering the history rather than relying only on cues visible near the decision point.


\bibliography{ref}  


\newpage

\appendix

\section{Technical Appendix}
\label{app:experimental_details}

This appendix collects the technical material that supports the main text. The subsections follow the paper narrative. They define the delayed-evidence setup for TRACE, specify the state representation, summarise training hyperparameters, state the signature and memory equations, define diagnostics, and give the adapter, variance, and case study details that are too long for the main paper.

\subsection{Problem Setup Details}
\label{app:problem_setup_details}

We consider demonstrations $\mathcal{D}=\{\tau_i\}_{i=1}^{N}$ with
\begin{equation}
    \tau_i=\{(o_t^i,s_t^i,a_t^i)\}_{t=1}^{T_i},
    \qquad
    x_t^i=(o_t^i,s_t^i),
    \qquad
    \mathcal{H}_t^i=(x_1^i,a_1^i,\ldots,x_{t-1}^i,a_{t-1}^i,x_t^i).
    \label{eq:app_demonstrations}
\end{equation}
Here $x_t$ is the current multi-view observation and robot state, and $\mathcal{H}_t$ is the causal history available before choosing the action at time $t$. A delayed-evidence task contains a past evidence variable $e_i=\eta(\mathcal{H}_{t_e}^i)$ observed in the history for some $t_e<t_b$. This variable can be the origin shelf, garment side, source tray, object identity, or another cue that determines a later branch. At a branch step $t=t_b$, the branch action $a_t$ is the action whose correct value depends on that earlier evidence.

The central ambiguity is that different past evidence can lead to nearly identical current observations at the branch point. There can be two trajectories with $e_i\neq e_j$ such that
\begin{equation}
    x_{t_b}^i\approx x_{t_b}^j,
    \qquad
    a_{t_b}^{i,\star}
    \neq
    a_{t_b}^{j,\star},
    \qquad
    p(a_{t_b}^{\star}\mid \mathcal{H}_{t_b}^i)
    \neq
    p(a_{t_b}^{\star}\mid \mathcal{H}_{t_b}^j).
    \label{eq:app_delayed_evidence}
\end{equation}
A policy of the form $\pi(a_t\mid x_t)$ must assign one action distribution to these ambiguous branch observations, so it is not sufficient for this setting. The needed object is a compact causal history summary $m_t=m(\mathcal{H}_t)$ such that $\pi(a_t\mid x_t,m_t)$ can recover the evidence-dependent branch decision while keeping the interface fixed size.

TRACE implements $m_t$ with an online signature-routed slot memory and the adapter condition derived from it. The past evidence variable $e_i$ is only an unobserved variable used to describe the problem. TRACE does not receive $e_i$ as a supervised label. It also does not receive branch labels, task identifiers, future observations, or future actions. During training and deployment, its memory update at time $t$ reads only the causal history contained in $\mathcal{H}_t$.

Each rollout is scored by decomposing the task into ordered subtasks and then into stages. This metric is inspired by partial-progress reporting in long-horizon manipulation benchmarks, including completed phases or subtasks in FurnitureBench and successful task-chain length in CALVIN-style evaluation~\citep{heo2025furniturebench,mees2022calvin}. It follows the reporting practice rather than reusing the same formula. For task $q$, the reported progress is
\begin{equation}
    \mathrm{Progress}_q
    =
    \frac{100}{N_qT_qS_q}
    \sum_{i=1}^{N_q}
    \sum_{k=1}^{T_q}
    \sum_{j=1}^{S_q}
    \mathbf{1}\!\left[
    \mathrm{stage}_{i,k,j}\ \mathrm{succeeds}
    \right].
    \label{eq:stage_progress}
\end{equation}
Here $N_q{=}25$ physical rollouts in the main evaluation, $T_q$ is the number of subtasks, and $S_q$ is the number of stages per subtask. Aggregate averages are task-balanced. Standard errors use rollout-level bootstrap resampling within each task.

\textbf{Baseline controls.}
\label{app:baseline_protocols}
Table~\ref{tab:baseline_protocols} gives the baseline protocol referenced in the experiments. The comparison keeps cameras, proprioception, action rate, train split normalisation, and held out evaluation lists matched across methods.

\begin{table}[H]
    \centering
    \caption{External VLA baseline protocol. Language conditioned models receive only a generic task prompt.}
    \label{tab:baseline_protocols}
    \setlength{\tabcolsep}{3.5pt}
    \renewcommand{\arraystretch}{1.05}
    \resizebox{\linewidth}{!}{%
    \scriptsize
    \begin{tabular}{@{}llll@{}}
        \toprule
        Baseline & Initialisation and trainable parts & Budget & Interface controls \\
        \midrule
        SmolVLA & \texttt{lerobot/smolvla\_base}, tuned adapter and state mapper, frozen vision & 180k updates & 3 RGB views, 17-D state, 50-step action chunks \\
        $\pi_{0.5}$ & \texttt{lerobot/pi05\_base} with the released policy adaptation path & 180k updates & 224px RGB views, 17-D state, 50-step action chunks \\
        X-VLA & \texttt{lerobot/xvla-base}, tuned policy transformer and soft prompts & 180k updates & Same views and state, 16-step action chunks, no cue prompt \\
        GR00T N1.6 & Released N1.6 adaptation with tuned embodiment and action head & 180k updates & Same rollout API, action rate, and held out episodes \\
        \bottomrule
    \end{tabular}}
\end{table}

The baseline protocol table is included to make the comparison auditable rather than to add another performance claim. The important control is that each baseline sees the same camera stream, proprioception layout, action rate, and held-out evaluation list. Language-conditioned methods receive a generic task prompt, so the delayed cue is not leaked through text.

\subsection{Model Size and Hyperparameters}
\label{app:model_size_hparams}

Parameter accounting and training hyperparameters are reported to support reproducibility and capacity fairness across policy families. They are not used as standalone performance claims. Table~\ref{tab:model_size_summary} separates total, trainable, and frozen capacity from the local checkpoint and model-cache audit. Table~\ref{tab:training_hparams} summarises the training schedules and optimiser settings, while the architectural settings that determine the TRACE module size are listed in Table~\ref{tab:prism_hparams}.

\begin{table}[H]
    \centering
    \caption{Model size summary for baselines and TRACE modules from the local checkpoint and model-cache audit.}
    \label{tab:model_size_summary}
    \setlength{\tabcolsep}{2.6pt}
    \renewcommand{\arraystretch}{1.08}
    \resizebox{\linewidth}{!}{%
    \scriptsize
    \begin{tabular}{@{}>{\raggedright\arraybackslash}p{0.14\linewidth}
                    >{\raggedright\arraybackslash}p{0.18\linewidth}
                    >{\centering\arraybackslash}p{0.13\linewidth}
                    >{\centering\arraybackslash}p{0.14\linewidth}
                    >{\raggedright\arraybackslash}p{0.22\linewidth}
                    >{\raggedright\arraybackslash}p{0.19\linewidth}@{}}
        \toprule
        Model or module & Count scope & Total parameters & Trainable parameters & Frozen parameters or frozen parts & Notes \\
        \midrule
        ACT & Full policy used as the regression base expert & $51.62$M & $51.62$M & $0$ & Same observations, state, action rate, and train split as TRACE Regression runs \\
        Diffusion Policy & Full policy used as the diffusion base expert & $270.96$M & $270.96$M & $0$ & Same observations, state, action rate, and train split as TRACE Diffusion runs \\
        SmolVLA & Fine tuned LeRobot baseline & $450.05$M & $99.88$M & $350.17$M VLM frozen & Uses the released fine tuning head and matched single task data \\
        $\pi_{0.5}$ & Fine tuned LeRobot baseline & $3.62$B & $693.42$M & $2.92$B PaliGemma VLM frozen & Uses the released adaptation path and matched single task data \\
        X-VLA & Fine tuned VLA baseline & $879.74$M & $879.74$M & $0$ & Tuned VLM, policy transformer, and soft prompts, with no cue prompt \\
        GR00T N1.6 & Fine tuned humanoid policy baseline & $2.72$B & $1.07$B & $1.66$B language/vision frozen & Tuned embodiment interface and action head under the matched rollout API \\
        TRACE updater & Shared signature routed memory updater only & $11.91$M & $11.91$M & $0$ & Excludes the base expert and policy specific adapter \\
        Regression adapter & TRACE adapter for the regression base expert & $0.79$M & $0.79$M & $0$ & Excludes the shared updater and the base expert \\
        Diffusion adapter & TRACE adapter for the diffusion base expert & $16.03$M & $16.03$M & $0$ & Excludes the shared updater and the base diffusion expert \\
        \bottomrule
    \end{tabular}}
\end{table}

The audit separates model capacity from the memory module. ACT and Diffusion Policy are the base experts for the two TRACE interfaces, while the VLA baselines include their released adaptation paths and frozen components. The TRACE updater is shared across adapters, so the added capacity can be attributed to a fixed memory module plus a small policy-facing translator rather than to a new action backbone.

Table~\ref{tab:prism_added_params} reports the additive parameter budget of TRACE for two policy-facing interfaces. The reference base expert is used only as the audited denominator for the percentage. The row name describes the TRACE interface rather than a fixed combined method. The added count includes the shared updater and the policy-specific adapter. The percentage is computed as $100\times N_{\mathrm{added}}/N_{\mathrm{base}}$ from the audited counts.

\begin{table}[H]
    \centering
    \caption{Additive TRACE parameter budget by policy-facing interface.}
    \label{tab:prism_added_params}
    \setlength{\tabcolsep}{2.8pt}
    \renewcommand{\arraystretch}{1.08}
    \resizebox{\linewidth}{!}{%
    \scriptsize
    \begin{tabular}{@{}>{\raggedright\arraybackslash}p{0.17\linewidth}
                    >{\raggedright\arraybackslash}p{0.18\linewidth}
                    >{\raggedright\arraybackslash}p{0.20\linewidth}
                    >{\centering\arraybackslash}p{0.13\linewidth}
                    >{\centering\arraybackslash}p{0.13\linewidth}
                    >{\centering\arraybackslash}p{0.10\linewidth}
                    >{\raggedright\arraybackslash}p{0.09\linewidth}@{}}
        \toprule
        TRACE interface & Reference base expert & Added TRACE parts & Base parameters & Added parameters & Increase & Notes \\
        \midrule
        Regression-policy interface & ACT audit checkpoint & Shared updater plus regression-interface adapter & $51.62$M & $12.69$M & $24.6\%$ & Interface only \\
        Diffusion-policy interface & Diffusion Policy audit checkpoint & Shared updater plus diffusion-interface adapter & $270.96$M & $27.94$M & $10.3\%$ & Interface only \\
        \bottomrule
    \end{tabular}}
\end{table}

The added-parameter table shows that the two interfaces have different capacity costs because they condition different base experts. The regression interface adds $12.69$M parameters over the audited ACT checkpoint, and the diffusion interface adds $27.94$M over the audited Diffusion Policy checkpoint. These counts are used for capacity accounting. They are not used as a substitute for the physical rollout comparisons in Table~\ref{tab:main_results}.

\begin{table}[H]
    \centering
    \caption{Training hyperparameters used by the reported protocol. All policies are trained for $180$k updates; batch sizes and action windows follow each policy family.}
    \label{tab:training_hparams}
    \setlength{\tabcolsep}{2.4pt}
    \renewcommand{\arraystretch}{1.08}
    \resizebox{\linewidth}{!}{%
    \scriptsize
    \begin{tabular}{@{}>{\raggedright\arraybackslash}p{0.12\linewidth}
                    >{\raggedright\arraybackslash}p{0.14\linewidth}
                    >{\raggedright\arraybackslash}p{0.14\linewidth}
                    >{\raggedright\arraybackslash}p{0.16\linewidth}
                    >{\raggedright\arraybackslash}p{0.20\linewidth}
                    >{\raggedright\arraybackslash}p{0.24\linewidth}@{}}
        \toprule
        Policy family & Code policy & Updates / batch & Action window & Optimiser and scheduler & Fine-tuning settings \\
        \midrule
        ACT regression & \texttt{act} & $180$k updates; batch $8$ or $32$ & Chunk $100$, execute $100$ & LeRobot ACT optimiser preset; AMP enabled & No TRACE memory; same cameras, state, action normalisation, train split, and evaluation API as TRACE Regression \\
        Diffusion Policy & \texttt{diffusion} & $180$k updates; batch $8$ & $2$ obs. steps, horizon $104$, execute $100$, drop last $3$ frames & LeRobot Diffusion optimiser preset; AMP enabled & No TRACE memory; same observation and action interface as TRACE Diffusion \\
        SmolVLA & \texttt{smolvla} & $180$k updates; batch $8$ or $16$ & $1$ obs. step, chunk $50$, execute $50$, $10$ flow steps & AdamW, lr $10^{-4}$, weight decay $10^{-10}$, grad clip $10$; $1000$ warmup steps and cosine decay over $30$k steps to $2.5{\times}10^{-6}$ & Frozen vision encoder; train action expert and state projection; generic task prompt only \\
        $\pi_{0.5}$ & \texttt{pi05} & $180$k updates; batch $1$ & $1$ obs. step, chunk $50$, execute $50$, $10$ inference steps & AdamW, lr $2.5{\times}10^{-5}$, weight decay $0.01$, grad clip $1$; $1000$ warmup steps and cosine decay over $30$k steps to $2.5{\times}10^{-6}$ & Frozen vision encoder; train action expert branch; mean/std state and action normalisation for the Zeno datasets \\
        X-VLA & \texttt{xvla} & $180$k updates; batch $1$ & $1$ obs. step, chunk $16$, execute $16$, $10$ denoising steps & AdamW, lr $10^{-4}$, weight decay $0$, grad clip $10$; $1000$ warmup steps and cosine decay over $30$k steps to $2.5{\times}10^{-6}$ & Train policy transformer and soft prompts; no cue prompt \\
        GR00T N1.6 & External adaptation & $180$k updates & Same rollout API and action rate as the robot policies & Released N1.6 adaptation recipe & Tune embodiment interface and action head; language and vision components frozen as in the parameter audit \\
        \bottomrule
    \end{tabular}}
\end{table}

The training table is intentionally separate from the architecture table below. Table~\ref{tab:training_hparams} records the optimisation budget, action horizon, and fine-tuning choices used in the reported training protocol. Table~\ref{tab:prism_hparams} records the memory architecture values that determine the TRACE module size and routing interface.

\subsection{Dataset and Demonstration Details}
\label{app:dataset_demonstration_details}

Table~\ref{tab:dataset_demonstration_details} makes the data accounting explicit for the five real robot tasks. The physical rollout count is the main evaluation count used in Equation~\ref{eq:stage_progress}. Tool, Book, and Laundry counts and horizons are audited from the processed LeRobot metadata. Cable and Medicine counts and horizons are audited from the local ROS bag timestamp scan using the same $30$ Hz sampling rule as the converter.

\begin{table}[H]
    \centering
    \caption{Dataset scale and SS details for the real robot evaluation protocol reported in this paper.}
    \label{tab:dataset_demonstration_details}
    \setlength{\tabcolsep}{2.4pt}
    \renewcommand{\arraystretch}{1.08}
    \resizebox{\linewidth}{!}{%
    \scriptsize
    \begin{tabular}{@{}>{\raggedright\arraybackslash}p{0.09\linewidth}
                    >{\centering\arraybackslash}p{0.10\linewidth}
                    >{\centering\arraybackslash}p{0.11\linewidth}
                    >{\centering\arraybackslash}p{0.12\linewidth}
                    >{\centering\arraybackslash}p{0.13\linewidth}
                    >{\centering\arraybackslash}p{0.12\linewidth}
                    >{\raggedright\arraybackslash}p{0.33\linewidth}@{}}
        \toprule
        Task & Train demos & Validation demos & Physical eval. rollouts & Episode length or avg. horizon & Subtasks / stages & Held-out factors \\
        \midrule
        Tool & 100 & 0 & 25 & $2{,}910$ frames ($97.0$s) & $2/4$ & Object instances, cue assignments, poses, lighting, distractors \\
        Book & 150 & 0 & 25 & $1{,}392$ frames ($46.4$s) & $3/4$ & Origin cue assignments, poses, routes, lighting, distractors \\
        Laundry & 60 & 0 & 25 & $2{,}220$ frames ($74.0$s) & $2/4$ & Garment instances, origin side assignments, poses, lighting, distractors \\
        Cable & 30 & 0 & 25 & $1{,}770$ frames ($59.0$s) & $1/4$ & Origin side assignments, device poses, lighting, distractors \\
        Medicine & 60 & 0 & 25 & $1{,}759$ frames ($58.6$s) & $2/4$ & Tray origin assignments, box poses, lighting, distractors, bedside props \\
        \bottomrule
    \end{tabular}}
\end{table}

The dataset table makes the evidence source explicit. Every task uses $25$ physical evaluation rollouts, and the reported progress scores in the main table are computed from those rollouts. The train demonstration counts and horizons come from audited metadata or ROS bag timestamp scans. The held-out factors show that the evaluation changes object instances, cue assignments, poses, lighting, and distractors rather than repeating the training episodes.

\subsection{State Representation and Normalisation}
\label{app:state_representation}
\label{app:state_hparams}

The policy backbones and TRACE use one overhead RGB camera, two wrist RGB cameras, and a 17-D proprioceptive state. The state is ordered as base planar velocity, base yaw velocity, left arm joint positions, and right arm joint positions. Fixed-base tasks keep the same layout and set the base channels to zero before signature computation. The zero mask is applied before train-split standardisation, so constant channels contribute no path increments.

At each control step, the masked state is standardised and appended to the streamed history path. The first observed state is used as the basepoint. TRACE omits the scalar signature coordinate. For $d_s{=}17$ and $p{=}3$, this gives $d_{\mathrm{sig}}=17+17^2+17^3=5{,}219$. Path signatures and one-step delta signatures are normalised with their own train-split statistics before the learned maps $\phi_g$ and $\phi_\Delta$. No cue label, language token, gripper command history, future action, or branch identifier is appended to the signature state.

\begin{table}[H]
    \centering
    \caption{Robot state channels used by TRACE signatures.}
    \label{tab:state_signature_details}
    \setlength{\tabcolsep}{3.2pt}
    \renewcommand{\arraystretch}{1.05}
    \resizebox{\linewidth}{!}{%
    \scriptsize
    \begin{tabular}{@{}lllll@{}}
        \toprule
        Channel block & Dim. & Units & Normalisation for signature input & Zeroed channel rule \\
        \midrule
        Base planar velocity $(v_x,v_y)$ & 2 & m/s & Train mean/std after the zero mask with std floor $10^{-6}$ & Fixed-base tasks set both channels to 0 \\
        Base yaw velocity $\omega$ & 1 & rad/s & Train mean/std after the zero mask with std floor $10^{-6}$ & Fixed-base tasks set this channel to 0 \\
        Left arm joints $q^{L}_{0,\ldots,6}$ & 7 & rad & Train mean/std, then append to the history path & Never zeroed \\
        Right arm joints $q^{R}_{0,\ldots,6}$ & 7 & rad & Train mean/std, then append to the history path & Never zeroed \\
        Full state path $\tilde s_{1,\ldots,t}$ & 17 & normalised units & Piecewise linear path over masked, standardised states & Constant zero channels add no increments \\
        Path signature $\xi_t$ & 5,219 & signature units & Train mean/std over history signatures, then $g_t=\phi_g(\xi_t)$ & Scalar coordinate omitted \\
        Delta signature $\delta_t$ & 5,219 & signature units & Train mean/std over one-step differences, then $\Delta g_t=\phi_\Delta(\delta_t)$ & $\delta_1=\mathbf{0}$ \\
        \bottomrule
    \end{tabular}}
\end{table}

The state table clarifies what can and cannot enter the signature. The routing key uses only the masked and standardised robot state path, with constant fixed-base channels contributing no path increments. No cue label, branch identifier, future action, or language token is appended. This keeps the memory condition causal and prevents the signature from becoming a hidden branch cue.

Table~\ref{tab:prism_hparams} lists the shared TRACE settings used in the reported runs. These values define the signature interface, memory width, and adapter dimensionality. The downstream policy losses and action heads are unchanged.

\begin{table}[H]
    \centering
    \caption{Core TRACE hyperparameters.}
    \label{tab:prism_hparams}
    \setlength{\tabcolsep}{4pt}
    \renewcommand{\arraystretch}{1.05}
    \resizebox{\linewidth}{!}{%
    \scriptsize
    \begin{tabular}{@{}lll@{}}
        \toprule
        Quantity & Reported setting & Notes \\
        \midrule
        Signature depth $p$ & 3 & Standard streamed signature \\
        Raw signature dim. $d_{\mathrm{sig}}$ & 5,219 & $17+17^2+17^3$, scalar term omitted \\
        Signature embedding dims. & 512 for $g_t$, 512 for $\Delta g_t$ & Shared by both adapters \\
        Slot count $K$ & 4 or 6 & Chosen by task horizon and branch diversity \\
        Slot width $d$ & 512 & Matches policy conditioning width \\
        History budget $L$ & 24 or 32 & Used for training-time causal history reconstruction \\
        History stride $r$ & 4 to 8 & Covers the recent tail at 30 FPS data rate \\
        Routing hidden size & 512 & Used for route query and key maps \\
        Adapter hidden size & 512 & Used by regression memory tokens and diffusion conditioning maps \\
        \bottomrule
    \end{tabular}}
\end{table}

These hyperparameters keep the shared TRACE updater identical across the two policy families. Depth $3$ signatures provide the routed history key, $512$-D memory slots match the policy conditioning width, and the fixed history budget controls the supervised training scan. The table also separates these memory settings from the base policy loss and action decoder, which remain unchanged.

\subsection{Signature Depth and Invariance}
\label{app:signature_depth}

Let $S_t$ be the piecewise-linear interpolation of the causal state history up to time $t$. TRACE uses the truncated path signature and its first difference
\begin{equation}
    \xi_t =
    \operatorname{Sig}_{\leq p}(S_t)
    \in\mathbb{R}^{d_{\mathrm{sig}}},
    \qquad
    \delta_t=\xi_t-\xi_{t-1},
    \qquad
    g_t=\phi_g(\xi_t),
    \quad
    \Delta g_t=\phi_{\Delta}(\delta_t).
    \label{eq:delta_signature}
\end{equation}
The initial delta is $\delta_1=\mathbf{0}$. The learned maps turn the deterministic signature vectors into routing features.

For an increasing time change $\alpha$ and a constant state offset $b$, the routing key uses
\begin{equation}
    \operatorname{Sig}_{\leq p}(S\circ\alpha)
    =
    \operatorname{Sig}_{\leq p}(S),
    \qquad
    \operatorname{Sig}^{S_0+b}_{\leq p}(S+b)
    =
    \operatorname{Sig}^{S_0}_{\leq p}(S).
    \label{eq:signature_routing_invariance}
\end{equation}
The second identity uses the basepointed signature. The address therefore depends on executed history geometry rather than sampling rate, execution speed, or a constant coordinate shift. The address remains order sensitive, so reversing the causal order changes the key.

For a state path with dimension $d_s$ and standard truncated signatures with the scalar coordinate omitted,
\begin{equation}
    d_{\mathrm{sig}}(p)=\sum_{k=1}^{p} d_s^k .
    \label{eq:signature_dim}
\end{equation}
The dimension grows exponentially with depth. With the 17-D state path used in our experiments, the jump from $p{=}3$ to $p{=}4$ increases the raw signature from $5{,}219$ to $88{,}740$ coordinates before learned compression.

\begin{table}[H]
    \centering
    \caption{Signature depth scaling for $d_s{=}17$.}
    \label{tab:signature_depth_ablation}
    \setlength{\tabcolsep}{3.5pt}
    \renewcommand{\arraystretch}{1.05}
    \resizebox{\linewidth}{!}{%
    \scriptsize
    \begin{tabular}{@{}lcccl@{}}
        \toprule
        Depth $p$ & Standard $d_{\mathrm{sig}}$ & Log signature dim. & Used in reported runs & Practical interpretation \\
        \midrule
        1 & 17 & 17 & No & Captures net displacement only \\
        2 & 306 & 153 & No & Adds pairwise order terms at low cost \\
        3 & 5,219 & 1,785 & Yes & Captures route and phase interactions while remaining practical for learned compression \\
        4 & 88,740 & 22,593 & No & Requires aggressive compression before routing \\
        \bottomrule
    \end{tabular}}
\end{table}

We use $p{=}3$ as a practical balance. Depth $1$ is close to a displacement descriptor and loses much of the ordering that distinguishes delayed branches. Depth $2$ is cheaper but may underrepresent multi-stage histories such as cue, transit, and branch setup. Depth $4$ substantially increases feature storage and learned-compression cost, making compression the dominant TRACE term.

Log signatures are attractive because they remove algebraic redundancies and reduce the feature dimension, as shown in Table~\ref{tab:signature_depth_ablation}. They are also less redundant as inputs to a learned map. The tradeoff lies in engineering and numerical complexity. Streamed online updates, delta features, normalisation statistics, and GPU support must match the deployment path. We therefore use standard streamed signatures in all reported experiments. Log signatures are an alternative focused on efficiency and remain outside the scope of these results.

\subsection{Diagnostic Metric Definitions}
\label{app:diagnostic_metrics}

For each held-out online evaluation rollout $e$ from task $q$ and history transform $T$, we run a causal diagnostic pass along the transformed history and compare it with the identity pass on the same online rollout. Let $I$ denote the identity transform, $t_b(e)$ the first ambiguous branch point, and $\mathcal{P}_e=\{t:t<t_b(e)\}$ the history indices before the branch point. The diagnostic inputs are the slot routing distributions $\omega_t^{T,e}\in\Delta^{K-1}$ for $t\in\mathcal{P}_e$, the branch point readout $z_{t_b(e)}^{T,e}\in\mathbb{R}^{d}$, and the branch action label $\hat b^{T,e}$ induced by the action head. The transform is applied only to the robot state path that produces TRACE signatures. The rollout identity, branch frame, branch label, and visual observations used by the policy come from the online execution and are unchanged, so the comparison isolates the effect of the transformed history route while preserving the online-rollout data source.

Route similarity measures whether the transformed history writes through the same memory slots as the identity online rollout pass before the branch decision. For a task $q$ with held out rollouts $\mathcal{E}_q$, the raw route score is
\begin{align}
    r(e,T)
    &=
    |\mathcal{P}_e|^{-1}
    \sum_{t\in\mathcal{P}_e}
    \frac{
    \langle \omega_t^{T,e},\omega_t^{I,e}\rangle
    }{
    \|\omega_t^{T,e}\|_2\,
    \|\omega_t^{I,e}\|_2
    },
    \\
    R_q(T)
    &=
    |\mathcal{E}_q|^{-1}
    \sum_{e\in\mathcal{E}_q} r(e,T).
    \label{eq:route_raw}
\end{align}
The routing vectors are nonnegative probability distributions, so the cosine similarity lies in $[0,1]$. High route similarity means that the transformed history follows the same sequence of soft slot addresses as the original Full TRACE rollout. The reported value is normalised to the Full TRACE identity online rollout pass on the same task,
\begin{equation}
    \operatorname{RouteSim}_q(T)
    =
    100\,
    \frac{R_q(T)}{R_q(I)}.
    \label{eq:route_norm}
\end{equation}
For the Full TRACE reference, $R_q(I)=1$ by construction. We keep the denominator explicit because all task level scores are reported relative to this reference before averaging across tasks.

Branch consistency measures whether the branch point evidence read from memory is preserved and whether it supports the same delayed branch action. We compute a continuous readout term and a discrete action agreement term,
\begin{align}
    C_q^{\mathrm{read}}(T)
    &=
    |\mathcal{E}_q|^{-1}
    \sum_{e\in\mathcal{E}_q}
    \frac{1+
    \frac{
    \langle z_{t_b(e)}^{T,e},z_{t_b(e)}^{I,e}\rangle
    }{
    \|z_{t_b(e)}^{T,e}\|_2\,
    \|z_{t_b(e)}^{I,e}\|_2
    }}{2},
    \\
    C_q^{\mathrm{act}}(T)
    &=
    |\mathcal{E}_q|^{-1}
    \sum_{e\in\mathcal{E}_q}
    \mathbf{1}\!\left[\hat b^{T,e}=\hat b^{I,e}\right],
    \\
    B_q(T)
    &=
    \frac{1}{2}
    \left(
    C_q^{\mathrm{read}}(T)+C_q^{\mathrm{act}}(T)
    \right),
    \\
    \operatorname{BranchCons}_q(T)
    &=
    100\,
    \frac{B_q(T)}{B_q(I)}.
    \label{eq:branch_norm}
\end{align}
The readout cosine is mapped from $[-1,1]$ to $[0,1]$, and the action term is the agreement rate with the original Full TRACE branch action. High branch consistency means that the transformed history exposes a similar memory readout at the first ambiguous decision and leads the action head to choose the same branch. The final table values are task balanced averages, $|\mathcal{Q}|^{-1}\sum_q\operatorname{Metric}_q(T)$. Standard errors are computed by bootstrap resampling held out rollouts within each task and recomputing the task balanced average.

Order reversal is used as a negative control because it preserves the held out episode, the branch frame, and many marginal state values, but it destroys the causal order of the history. This is exactly the information that path signatures and the slot routing keys are meant to encode. Order-preserving transforms such as time resampling or speed jitter keep both diagnostics high, while reversal reduces them when TRACE uses ordered history evidence rather than cues visible only near the decision point or the same states without their order. Full TRACE is the identity online rollout reference, so both normalised diagnostics are reported as $100$ for $I$ by construction. These quantities are computed only after training. They are not losses, they are not used for model selection, and no gradient is taken through them.

\subsection{Long-Horizon Memory Stability}
\label{app:stability}

The invariance diagnostics above test whether the branch decision depends on ordered history evidence. We also run a long-history stability pass to check whether the fixed slot memory remains usable as held-out online rollouts provide longer causal histories before the branch readout. This pass is a diagnostic of the reported horizons, not a theoretical guarantee for arbitrary episode length.

The protocol has two forms. In the reported-horizon online-history pass, each held-out physical rollout supplies the causal history. The trained updater is applied along that executed history, and the memory state is recorded at the first ambiguous branch point. In extended-history diagnostics, we continue the same online-history accumulation with measured causal segments from recorded online rollouts and deployed diagnostic passes. The extensions use repeated distractor segments, speed jitter, sparse resampling, and additional shared-route online segments when those segments are available in the recorded diagnostic source. The branch frame, branch label, visual observations, and policy weights are held fixed, so changes in the diagnostic scores reflect changes in the memory route and stored history evidence. These diagnostic rows are not extrapolated rollout success rates.

For long-history stability, we record slot churn, write entropy, slot occupancy, branch readout consistency, and branch decision consistency. For an episode with write distributions $\omega_1,\ldots,\omega_T$, these are
\begin{align}
    \mathrm{Churn}
    &=
    (T-1)^{-1}\sum_{t=2}^{T}
    \mathbf{1}\!\left[
    \arg\max_j \omega_t^{(j)}
    \neq
    \arg\max_j \omega_{t-1}^{(j)}
    \right],
    \\
    \mathrm{WriteEnt}
    &=
    (T\log K)^{-1}
    \sum_{t=1}^{T}
    \left(
    -\sum_{j=1}^{K}\omega_t^{(j)}\log\omega_t^{(j)}
    \right),
    \\
    \mathrm{Occupancy}
    &=
    K^{-1}
    \sum_{j=1}^{K}
    \mathbf{1}\!\left[
    \exists t\leq T,\,
    j=\arg\max_{j'}\omega_t^{(j')}
    \right].
    \label{eq:stability_metrics}
\end{align}
Branch readout consistency is the cosine similarity between the branch readout from the extended history and the readout from the matched reported-horizon online-history pass. Branch decision consistency is the corresponding action agreement rate.

\begin{table}[H]
    \centering
    \caption{Long-horizon memory stability diagnostics for fixed slot TRACE memory. Values are task-balanced means over held-out rollouts. Readout and decision consistency compare each extension with the matched reported-horizon branch readout.}
    \label{tab:long_horizon_stability}
    \setlength{\tabcolsep}{3.0pt}
    \renewcommand{\arraystretch}{1.05}
    \resizebox{\linewidth}{!}{%
    \scriptsize
    \begin{tabular}{@{}lcccccc@{}}
        \toprule
        Diagnostic pass & History extension & Slot churn$\downarrow$ & Write entropy$\downarrow$ & Slot occupancy$\uparrow$ & Readout cons.$\uparrow$ & Decision cons.$\uparrow$ \\
        \midrule
        Reported-horizon online history & $1.0\times$ & $0.010$ & $0.356$ & $0.646$ & $1.000$ & $1.000$ \\
        Extended history & $1.25\times$ & $0.017$ & $0.374$ & $0.690$ & $0.982$ & $0.968$ \\
        Extended history & $1.5\times$ & $0.025$ & $0.397$ & $0.710$ & $0.962$ & $0.936$ \\
        Extended history & $2.0\times$ & $0.041$ & $0.431$ & $0.744$ & $0.928$ & $0.904$ \\
        Repeated distractor extension & $1.5\times$ & $0.052$ & $0.462$ & $0.758$ & $0.907$ & $0.872$ \\
        Shared-route online extension & $2.0\times$ & $0.035$ & $0.412$ & $0.737$ & $0.943$ & $0.916$ \\
        \bottomrule
    \end{tabular}}
\end{table}

The stability readout is a measured diagnostic over the evaluated extensions rather than a proof for arbitrary horizons. Across the evaluated extensions, slot churn stays near $0.05$ or lower, occupancy remains broad but noncollapsed, and branch decision consistency remains at least $0.872$ even under repeated distractor segments. The larger entropy under the distractor extension comes from the added segment's reuse of ambiguous shared-route observations, while the branch readout remains close to the reported-horizon reference. Rising churn or entropy together with falling branch consistency is the overwrite or diffuse-write failure signature. The diagnostic therefore supports the memory claims only over the evaluated horizons and history extensions. It does not show that fixed slots avoid overwriting old information for arbitrarily long rollouts.

\subsection{Adapter Architectures}
\label{app:adapter_architectures}
\label{app:runtime_latency}

Both TRACE variants use the same signature-indexed memory updater. The main text names the two variants by their base objectives, Regression and Diffusion. In this architecture subsection, the same variants are described as the regression adapter and diffusion adapter because the figure and table refer to the policy-facing modules that present the memory readout to each base policy. For a policy family indexed by $e$,
\begin{equation}
    H_t=(M_t,z_t^{\mathrm{mem}},g_t,\Delta g_t),
    \qquad
    \mathcal{A}^{(e)}_{\theta}:\mathcal{H}_{\mathrm{TRACE}}\rightarrow\mathcal{C}_e,
    \qquad
    c_t^{(e)}=\mathcal{A}^{(e)}_{\theta}(H_t).
    \label{eq:adapter_interface}
\end{equation}
Here $\mathcal{C}_e$ is the native conditioning space of the downstream policy family. The adapter changes the policy input condition, but it leaves the action parameterisation and imitation objective unchanged.

\begin{table}[H]
    \centering
    \caption{Adapter interfaces for the regression and diffusion backbones.}
    \label{tab:adapter_architectures}
    \setlength{\tabcolsep}{3.0pt}
    \renewcommand{\arraystretch}{1.08}
    \resizebox{\linewidth}{!}{%
    \scriptsize
    \begin{tabular}{@{}>{\raggedright\arraybackslash}p{0.13\linewidth}
                    >{\raggedright\arraybackslash}p{0.22\linewidth}
                    >{\raggedright\arraybackslash}p{0.25\linewidth}
                    >{\raggedright\arraybackslash}p{0.27\linewidth}
                    >{\raggedright\arraybackslash}p{0.16\linewidth}@{}}
        \toprule
        Adapter & TRACE inputs & Adapter computation & Expert conditioning produced & Expert loss \\
        \midrule
        Regression adapter & Slot memory $M_t\in\mathbb{R}^{K\times512}$, readout $z_t^{\mathrm{mem}}$, signature embeddings $g_t,\Delta g_t$ & Map each slot with $W_M$. Map $[z_t^{\mathrm{mem}},g_t,\Delta g_t]$ into one summary token. Optionally use the summary for feature modulation & 512-D memory tokens concatenated to the regression policy attention memory. The action head regresses the native action chunk & Unchanged chunk L1 loss plus the standard optional KL term \\
        Diffusion adapter & Same $M_t,z_t^{\mathrm{mem}},g_t,\Delta g_t$ from the shared updater & Pool slots with readout attention weights. Concatenate $\operatorname{pool}(M_t)$, $z_t^{\mathrm{mem}}$, $g_t$, and $\Delta g_t$. Pass the result through a zero-initialised MLP & 512-D additive global conditioning vector appended to time-step and action conditioning. Diffusion denoising iterations are unchanged & Unchanged diffusion denoising loss over the action trajectory \\
        Shared updater & Masked state path, pooled multi-view visual feature $v_t$, proprioceptive embedding $p_t$, and previous slots $M_{t-1}$ & Compute $g_t,\Delta g_t$. Route a write distribution $\omega_t$. Update gated slot candidates and read slots with the current visual-state query & Policy-agnostic TRACE memory state $H_t$ consumed by either adapter & Auxiliary balance, entropy, and consistency losses only during training \\
        \bottomrule
    \end{tabular}}
\end{table}

The adapter comparison shows where the two TRACE variants differ. Both consume the same routed memory state, but the regression adapter exposes it as attention memory tokens and the diffusion adapter exposes it as a global conditioning vector. This is why the main experiments can attribute shared gains to the causal memory while still allowing each policy family to keep its own action loss.

Table~\ref{tab:runtime_latency} separates the unchanged base policy forward pass from TRACE's per-step computation. Entries are mean / p95 latency. The regression row is measured over 1,900 post-warmup control steps from the streaming regression policy profile on an NVIDIA GeForce RTX 5090 with CUDA 12.8 and PyTorch 2.9.1. The diffusion row combines the measured shared TRACE path with recorded diffusion and adapter timings from the deployed TRACE-conditioned policy.

\begin{table}[H]
    \centering
    \caption{Per-step deployment latency breakdown in milliseconds.}
    \label{tab:runtime_latency}
    \setlength{\tabcolsep}{3.5pt}
    \renewcommand{\arraystretch}{1.05}
    \resizebox{\linewidth}{!}{%
    \scriptsize
    \begin{tabular}{@{}lcccccccc@{}}
        \toprule
        Policy path & Base forward & Signature & Slot update & Readout & Adapter & TRACE overhead & Total loop & Overhead \\
        \midrule
        Regression adapter & 5.90 / 7.21 & 0.52 / 0.69 & 0.90 / 1.26 & 0.58 / 0.74 & 0.10 / 0.15 & 2.11 / 2.66 & 8.32 / 10.02 & 35.8\% \\
        Diffusion adapter$^\dagger$ & 118.00 / 142.00 & 0.52 / 0.69 & 0.90 / 1.26 & 0.58 / 0.74 & 0.14 / 0.21 & 2.15 / 2.72 & 120.46 / 144.87 & 1.8\% \\
        \bottomrule
    \end{tabular}}
    \vspace{2pt}
    \footnotesize{$^\dagger$ Diffusion timing uses the shared measured TRACE path. Base forward is the 100-step diffusion denoising call.}
\end{table}

The latency table shows that TRACE adds a small fixed computation before the unchanged base policy call. For the regression adapter, the overhead is visible because the base policy is already fast. For the diffusion adapter, the same memory path is small compared with the 100-step diffusion denoising call. The table therefore supports the claim that TRACE is an online conditioning module rather than a second policy pass or an offline retrieval procedure.

\subsection{Memory Update, Training, and Online Inference}
\label{app:memory_update_losses}
\label{app:adapter_update}

\begin{figure}[!htbp]
    \centering
    \includegraphics[width=\linewidth]{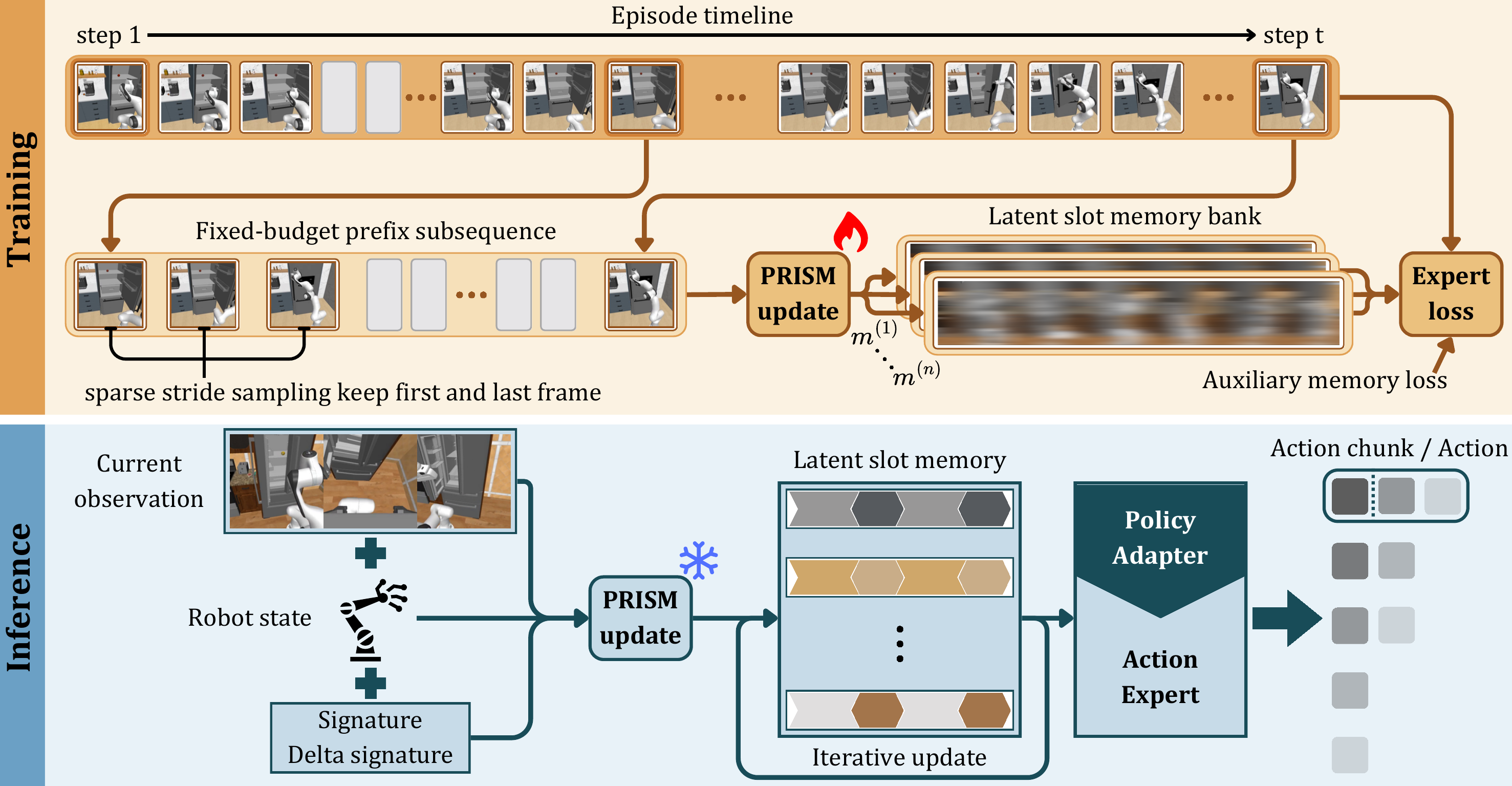}
    \caption{Training and inference consistency. Training scans the masked fixed-budget history available online, and deployment applies the same updater once per executed step.}
    \label{fig:prism_train_infer}
\end{figure}

At each step, camera features are pooled as $v_t=C^{-1}\sum_c\rho(f_{\mathrm{vis}}(o_t^{(c)}))$, and the proprioceptive embedding is $p_t=\phi_s(\tilde s_t)$. These features form the visual-state content $e_t=[v_t,p_t]$. The address feature $q_t$ is built from the projected path signature and delta signature. When first-frame anchoring is enabled, $q_t$ also includes the causal first-frame anchor. With this address vector, the route query and write distribution are
\begin{align}
    \rho_t &= \phi_{\rho}(q_t),
    \qquad
    \bar m_{t-1,k}=m_{t-1,k}+\lambda_{\eta}\eta_k,
    \\
    \ell_{t,k}
    &=
    \frac{(W_q\rho_t)^\top W_k\bar m_{t-1,k}}
    {\tau\sqrt{d_r}},
    \qquad
    \omega_{t,k}
    =
    \frac{\exp(\ell_{t,k})}{\sum_{j=1}^{K}\exp(\ell_{t,j})}.
    \label{eq:slot_route}
\end{align}
where $\eta_k$ is a fixed sinusoidal slot identity when enabled; setting $\lambda_{\eta}=0$ gives the adapters that do not use slot identity. The content proposal, write strength, and slot update are
\begin{align}
    u_t &= \phi_w(e_t,q_t),
    \\
    \tilde m_{t,k}
    &=
    \tanh\!\left(\phi_m(\bar m_{t-1,k},u_t,\rho_t)\right),
    \\
    \beta_{t,k}
    &=
    \omega_{t,k}
    \sigma\!\left(\phi_{\beta}(\bar m_{t-1,k},u_t,\rho_t)\right),
    \\
    m_{t,k}
    &=
    (1-\beta_{t,k})m_{t-1,k}
    +
    \beta_{t,k}\tilde m_{t,k}.
    \label{eq:slot_update}
\end{align}
The memory readout uses a separate attention distribution:
\begin{align}
    u_t^r
    &=
    \phi_r(e_t,\rho_t),
    \\
    \bar m_{t,k}
    &=
    m_{t,k}+\lambda_{\eta}\eta_k,
    \\
    a_{t,k}
    &=
    \frac{(u_t^r)^\top W_k^r\bar m_{t,k}}{\sqrt d},
    \\
    \alpha_{t,k}
    &=
    \frac{\exp(a_{t,k})}{\sum_{j=1}^{K}\exp(a_{t,j})},
    \\
    z_t^{\mathrm{mem}}
    &=
    \sum_{k=1}^{K}\alpha_{t,k}W_v^r m_{t,k}.
    \label{eq:slot_readout}
\end{align}

During training, each target time $t$ uses a padded fixed-budget causal history index sequence $I_{L,r}(t)=(i_1,\ldots,i_L)\subseteq\{1,\ldots,t\}$ with valid mask $b_l$. The selected valid entries contain the first available frame, the current frame, and a recent stride-sampled tail. Starting from $M^{(0)}=\mathbf{0}$, the scan is
\begin{equation}
    M^{(l)}
    =
    \mathcal{U}_{\theta}
    (M^{(l-1)},o_{i_l},s_{i_l},q_{i_l},b_l),
    \qquad
    l=1,\ldots,L.
    \label{eq:history_scan}
\end{equation}
This scan applies the same updater that deployment uses once per online step with the cached memory state. The final scanned memory forms $H_t$, and the native policy loss for expert family $e$ is
\begin{equation}
    \mathcal{L}_{\mathrm{policy}}^{(e)}
    =
    \mathbb{E}_{(\tau,t)\sim\mathcal{D}}
    \left[
    \ell_e\!\left(
    \mathcal{E}_{\psi}^{(e)}
    (x_t,\mathcal{A}^{(e)}_{\theta}(H_t)),
    y_t^\star
    \right)
    \right].
    \label{eq:native_policy_loss}
\end{equation}
The finite-slot updater has three common failure modes during training. Slot collapse writes most evidence through a few slots. Diffuse writing spreads one observation across many slots, making later addresses weak. Read-write mismatch stores information in a form that the readout cannot recover after subsequent updates. TRACE uses one lightweight stabiliser for each case, then keeps the downstream imitation objective unchanged.
\begin{equation}
    \mathcal{L}
    =
    \mathcal{L}_{\mathrm{policy}}^{(e)}
    +
    \lambda_{\mathrm{bal}}\mathcal{L}_{\mathrm{bal}}
    +
    \lambda_{\mathrm{ent}}\mathcal{L}_{\mathrm{ent}}
    +
    \lambda_{\mathrm{cons}}\mathcal{L}_{\mathrm{cons}}.
    \label{eq:total_loss}
\end{equation}
For the $l$-th selected history element of target $t$, let $\omega_{t,l}$ be the slot-routing distribution, $u_{t,l}$ the write proposal, and $z_{t,l}^{\mathrm{mem}}$ the memory readout. With $N_v=\sum_{t,l}b_{t,l}$ and $\bar\omega^{(j)}=N_v^{-1}\sum_{t,l}b_{t,l}\omega_{t,l}^{(j)}$, the auxiliary terms are
\begin{align}
    \mathcal{L}_{\mathrm{bal}}
    &=
    K^{-1}\sum_{j=1}^{K}
    \left(\bar\omega^{(j)}-K^{-1}\right)^2,
    \\
    \mathcal{L}_{\mathrm{ent}}
    &=
    (N_v\log K)^{-1}
    \sum_{t,l}b_{t,l}
    \left(
    -\sum_{j=1}^{K}
    \omega_{t,l}^{(j)}\log\omega_{t,l}^{(j)}
    \right),
    \\
    \mathcal{L}_{\mathrm{cons}}
    &=
    N_v^{-1}
    \sum_{t,l}b_{t,l}
    d^{-1}
    \left\|
    \tanh z_{t,l}^{\mathrm{mem}}
    -
    \tanh u_{t,l}
    \right\|_2^2 .
    \label{eq:memory_regularizers}
\end{align}
$\mathcal{L}_{\mathrm{bal}}$ penalises uneven average routing, so it discourages slot collapse. $\mathcal{L}_{\mathrm{ent}}$ penalises high-entropy routing at each valid write step, so it discourages diffuse writing. $\mathcal{L}_{\mathrm{cons}}$ aligns the bounded memory readout with the current write proposal, so it reduces read-write mismatch.

These terms are finite-capacity stabilisers rather than theoretical guarantees. They bias training toward balanced, addressable, and readable writes, but they do not prove that every piece of causal information remains recoverable. Since the memory has a fixed number of slots and a bounded slot width, sufficiently long horizons or trajectories with more pieces of information than the memory can represent may still force multiple facts to share capacity. In such regimes, later updates can overwrite or blur earlier content, especially when evidence is repeatedly routed to nearby slots or when visually similar observations require different interpretations.

The auxiliary terms act only on memory. The downstream policy loss remains chunk L1 with the standard optional KL term for Regression, and the native diffusion denoising loss for Diffusion.

\begin{algorithm}[H]
    \small
    \caption{Causal TRACE memory update and adapter conditioning.}
    \label{alg:online_update}
    \begin{algorithmic}[1]
        \Require Observation $o_t$, robot state $s_t$, previous memory $M_{t-1}$, previous signature $\xi_{t-1}$, optional first-frame anchor $a^0$, state mask $Z_q$, train-split normalisation statistics, policy family $e\in\{\mathrm{regression},\mathrm{diffusion}\}$.
        \Ensure Updated memory $M_t$, signature $\xi_t$, adapter conditioning $c_t$, and policy prediction $\hat y_t$.
        \State Apply the task mask and state normalisation.
        \[
            \bar s_t=Z_q(s_t),
            \qquad
            \tilde s_t=(\bar s_t-\mu_s)/(\sigma_s+\epsilon).
        \]
        \State Append $\tilde s_t$ to the piecewise linear history path $S_t$.
        \State Compute $\xi_t=\operatorname{Sig}_{\leq p}(S_t)$ and $\delta_t=\xi_t-\xi_{t-1}$.
        \State Remove the scalar signature coordinate, standardise $\xi_t$ and $\delta_t$, then map them to $g_t=\phi_g(\xi_t)$ and $\Delta g_t=\phi_\Delta(\delta_t)$.
        \State Encode the current observation as $v_t=C^{-1}\sum_c\rho(f_{\mathrm{vis}}(o_t^{(c)}))$ and $p_t=\phi_s(\tilde s_t)$.
        \State Compute address features $q_t=[g_t,\Delta g_t]$, appending $a^0$ only when first-frame anchor routing is enabled, and set $\rho_t=\phi_{\rho}(q_t)$.
        \State Route the write with Equation~\ref{eq:slot_route}.
        \State Form gated write candidates from visual-state content and address state, then update slots with Equation~\ref{eq:slot_update}.
        \State Read memory with Equation~\ref{eq:slot_readout}.
        \State Construct $H_t=(M_t,z_t^{\mathrm{mem}},g_t,\Delta g_t)$ and $c_t^{(e)}=\mathcal{A}_{\theta}^{(e)}(H_t)$.
        \State Predict $\hat y_t=f_\psi(z_t,c_t^{(e)})$ with the unchanged base policy.
    \end{algorithmic}
\end{algorithm}

\subsection{Additional Case Studies}
\label{app:additional_case_studies}

Table~\ref{tab:additional_case_studies} summarises the concrete delayed-evidence structure behind the five tasks. These case studies connect the diagnostic quantities in the main paper to rollout-level behaviour and give the evidence source for the baseline failure patterns discussed in the main text.

\begin{table}[H]
    \centering
    \caption{Additional task-level case studies linking baseline failures, evidence, and TRACE behaviour.}
    \label{tab:additional_case_studies}
    \setlength{\tabcolsep}{2.0pt}
    \renewcommand{\arraystretch}{1.10}
    \resizebox{\linewidth}{!}{%
    \scriptsize
    \begin{tabular}{@{}>{\raggedright\arraybackslash}p{0.09\linewidth}
                    >{\raggedright\arraybackslash}p{0.17\linewidth}
                    >{\raggedright\arraybackslash}p{0.17\linewidth}
                    >{\raggedright\arraybackslash}p{0.19\linewidth}
                    >{\raggedright\arraybackslash}p{0.19\linewidth}
                    >{\raggedright\arraybackslash}p{0.19\linewidth}@{}}
        \toprule
        Task & History evidence & Ambiguous branch & Baseline failure & Failure evidence & TRACE behaviour \\
        \midrule
        Tool & Initial object identity determines the later tool sequence & Object and tool are no longer jointly visible when the sequence must branch & Wrong tool order at the branch, or correct branch followed by contact loss & Table~\ref{tab:main_results} shows high Tool variance, and Table~\ref{tab:tool_failure_taxonomy} separates delayed-memory errors from contact losses & Stores the object-dependent history and improves branch selection, while contact-rich stages still create variance \\
        Book & Origin shelf determines route and placement & After pickup and transit, the overhead view is similar across origins & Places on a visually plausible target that is inconsistent with the origin route & Figure~\ref{fig:rollout_results_combined} shows similar views after transit, and Table~\ref{tab:memory_module_results} shows the matched memory gap & Reads the origin-dependent slot near placement and preserves route-specific evidence \\
        Laundry & Origin side determines brush-and-basket versus fold-and-store & Garment pose becomes visually similar after the shared carry segment & Executes the visually dominant routine regardless of origin side & Tables~\ref{tab:main_results} and~\ref{tab:memory_module_results} show the largest gains on this origin-side branch task & Routes the side cue during pickup and reuses it at the later branch \\
        Cable & Origin side determines the matched device & Traversal and local device pose provide evidence still visible near the device choice & Reaches the workspace but can choose the wrong device when local geometry is ambiguous & Tables~\ref{tab:main_results} and~\ref{tab:memory_module_results} show closer scores, consistent with partial evidence from current geometry & Memory helps at the device choice, but local manipulation still contributes many credited stages \\
        Medicine & Tray origin determines box selection and return & Boxes enter a shared bedside area before the return decision & Selects or returns the wrong box after a correct past pickup & Tables~\ref{tab:main_results} and~\ref{tab:memory_module_results} show gains on the tray-origin branch, and Table~\ref{tab:diagnostic_results} supports history-sensitive readout & Keeps the tray-origin cue available through the shared staging segment \\
        \bottomrule
    \end{tabular}}
\end{table}

The case studies explain why the same memory module has different task-level effects. Book, Laundry, and Medicine place much of the later score on remembering an origin cue, so TRACE changes the dominant branch error pattern. Cable still contains useful local geometry near the device choice, which narrows the gap, while Tool combines the delayed branch with contact-heavy execution and therefore needs the separate variance analysis in Appendix~\ref{app:variance_failure}.

\subsection{Variance and Failure Analysis}
\label{app:variance_failure}

The Tool task has the highest standard error in Table~\ref{tab:main_results}. Its progress score combines a discrete delayed memory decision with several contact-rich manipulation stages after the decision. A rollout can remember the correct object and tool branch but lose many credited stages due to grasp slip, weak tool contact, or recovery timeout. A wrong delayed branch can also remove several downstream stages at once. This explains why Tool is noisier than Book, Laundry, and Medicine, where the origin cue more directly determines the branch score.

Table~\ref{tab:leave_one_task_out_main} reports a simple robustness check for the main averages. The drop-Tool column removes the task with the largest SE and recomputes the task-balanced mean from the remaining four task means. TRACE Regression remains the strongest method at $72.92$, and TRACE Diffusion remains above the strongest non-TRACE baseline at $59.79$ versus $51.71$. Using only the reported task SEs gives drop-Tool 95\% intervals of $[71.28,74.55]$ for TRACE Regression, $[57.10,62.48]$ for TRACE Diffusion, and $[48.13,55.29]$ for the strongest non-TRACE baseline. This check uses no Tool rollouts, so the average gain is not carried by the high-variance task. The table is computed directly from Table~\ref{tab:main_results}.

\begin{table}[H]
    \centering
    \caption{Leave-one-task-out task-balanced averages for the main comparison. Values are recomputed from the task means in Table~\ref{tab:main_results}.}
    \label{tab:leave_one_task_out_main}
    \setlength{\tabcolsep}{2.4pt}
    \renewcommand{\arraystretch}{1.05}
    \resizebox{\linewidth}{!}{%
    \scriptsize
    \begin{tabular}{@{}lcccccc@{}}
        \toprule
        Method & All tasks & Drop Tool & Drop Book & Drop Laundry & Drop Cable & Drop Medicine \\
        \midrule
        Diffusion Policy & 25.00 & 26.75 & 28.17 & 25.75 & 22.75 & 21.58 \\
        ACT & 25.50 & 24.12 & 28.46 & 29.00 & 20.88 & 25.04 \\
        $\pi_{0.5}$ & 50.47 & 51.71 & 50.33 & 51.21 & 50.83 & 48.25 \\
        GR00T N1.6 & 30.60 & 28.50 & 35.08 & 32.25 & 28.75 & 28.42 \\
        SmolVLA & 28.33 & 29.17 & 30.42 & 32.42 & 22.92 & 26.75 \\
        X-VLA & 33.90 & 41.00 & 30.62 & 32.12 & 33.38 & 32.38 \\
        \midrule
        \textbf{Ours (Regression)} & \textbf{69.23} & \textbf{72.92} & \textbf{65.79} & \textbf{66.29} & \textbf{73.79} & \textbf{67.38} \\
        \textbf{Ours (Diffusion)} & 59.53 & 59.79 & 57.33 & 56.79 & 61.66 & 62.08 \\
        \bottomrule
    \end{tabular}}
\end{table}

Table~\ref{tab:tool_bootstrap_ci} separates the noisy Tool score from the aggregate conclusion. Tool has a wide interval for both TRACE variants because a discrete delayed branch is followed by several contact-rich stages. The same table also reports the drop-Tool averages and their gaps against the strongest non-TRACE baseline, using the reported task SEs from Table~\ref{tab:main_results}. Regression keeps a $21.21$ point drop-Tool advantage over $\pi_{0.5}$, and Diffusion keeps an $8.08$ point advantage.

\begin{table}[H]
    \centering
    \caption{Tool-task uncertainty and drop-Tool robustness for TRACE variants. Tool intervals use the reported mean $\pm 1.96$ SE. Drop-Tool intervals are computed from the reported task SEs in Table~\ref{tab:main_results}.}
    \label{tab:tool_bootstrap_ci}
    \setlength{\tabcolsep}{3.2pt}
    \renewcommand{\arraystretch}{1.08}
    \resizebox{\linewidth}{!}{%
    \scriptsize
    \begin{tabular}{@{}>{\raggedright\arraybackslash}p{0.24\linewidth}
                    >{\centering\arraybackslash}p{0.18\linewidth}
                    >{\centering\arraybackslash}p{0.18\linewidth}
                    >{\centering\arraybackslash}p{0.18\linewidth}
                    >{\centering\arraybackslash}p{0.18\linewidth}@{}}
        \toprule
        Method & Tool mean $\pm$ SE & Tool 95\% interval & Drop-Tool avg. & Drop-Tool gap vs. $\pi_{0.5}$ \\
        \midrule
        Ours (Regression) & \score{54.50}{17.96} & $[19.30,89.70]$ & $72.92\,[71.28,74.55]$ & $+21.21\,[17.27,25.15]$ \\
        Ours (Diffusion) & \score{58.50}{10.17} & $[38.57,78.43]$ & $59.79\,[57.10,62.48]$ & $+8.08\,[3.60,12.56]$ \\
        \bottomrule
    \end{tabular}}
\end{table}

Table~\ref{tab:tool_failure_taxonomy} gives the failure categories used to interpret the high Tool variance. The categories separate whether TRACE reaches the delayed branch with the correct object-conditioned decision from later contact-rich losses. This distinction matters because Tool can produce mid-range or low progress for different reasons. Some rollouts preserve the delayed cue but lose downstream stages through contact, while others fail the memory-dependent branch itself.

\begin{table}[H]
    \centering
    \caption{Tool failure taxonomy for interpreting the high-variance rollouts. The categories separate delayed-memory failures from contact and recovery failures after the branch.}
    \label{tab:tool_failure_taxonomy}
    \setlength{\tabcolsep}{2.8pt}
    \renewcommand{\arraystretch}{1.08}
    \resizebox{\linewidth}{!}{%
    \scriptsize
    \begin{tabular}{@{}>{\raggedright\arraybackslash}p{0.18\linewidth}
                    >{\raggedright\arraybackslash}p{0.34\linewidth}
                    >{\raggedright\arraybackslash}p{0.20\linewidth}
                    >{\raggedright\arraybackslash}p{0.20\linewidth}@{}}
        \toprule
        Failure class & Operational definition & Score pattern & Interpretation \\
        \midrule
        Correct branch with contact loss & The object-dependent tool order is selected correctly, then progress drops from grasp slip, weak contact, or incomplete tool engagement & Shared setup and branch stages receive credit, but later manipulation stages are missing & Memory succeeds, while execution noise lowers total progress \\
        Wrong branch or tool order & The shared setup reaches the delayed decision, but the selected tool sequence does not match the initial object identity & past stages receive credit, followed by a sharp loss at the branch and its downstream stages & Direct delayed-evidence error. This is the failure mode TRACE is designed to reduce \\
        Recovery timeout & Manipulation stalls during recovery after a partial success and exceeds the timeout & Progress plateaus after partial later-stage credit & Separates memory retention from long-horizon recovery and contact robustness \\
        Pre-branch setup failure & The rollout fails before the delayed decision can be evaluated & Low progress before the branch point & Does not test delayed memory, but still contributes to rollout-level Tool variance \\
        Other or unclassified & The score trace or video evidence does not cleanly match the categories above & Mixed or inconsistent stage evidence & Reserved for audit before making branch-level rate claims \\
        \bottomrule
    \end{tabular}}
\end{table}

\end{document}